%% file: main.tex
\documentclass[11pt]{article}
\usepackage{opendrivelab}
\newlength{\orglogoheight}
\title{World Engine: Towards the Era of Post-Training for Autonomous Driving}

\orglabel{}
\orglogo{%
  \makebox[\linewidth][l]{%
    \includegraphics[height=0.8cm,keepaspectratio]{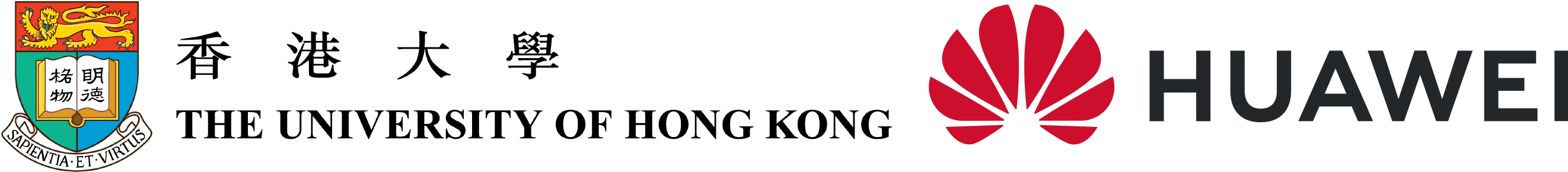}%
  }%
}

\contributors{%
  \small
  Tianyu Li\textsuperscript{4,3*}\quad 
  Li Chen\textsuperscript{4*}\quad 
  Caojun Wang\textsuperscript{3*}\quad 
  Haochen Liu\textsuperscript{7*}\quad 
  Kashyap Chitta\textsuperscript{5,6*}\quad 
  Zhenjie Yang\textsuperscript{1*}\quad \\
  Yuhang Lu\textsuperscript{1}\quad
  Naisheng Ye\textsuperscript{1}\quad
  Yihang Qiu\textsuperscript{1}\quad
  Yufei Wang\textsuperscript{2}\quad
  Luoxi Zou\textsuperscript{2}\quad
  Jiaxin Peng\textsuperscript{2}\quad
  Jin Pan\textsuperscript{2}\quad
  Zhaoyu Su\textsuperscript{2}\\
  Andrei Bursuc\textsuperscript{8}\quad
  Shengbo Eben Li\textsuperscript{9}\quad
  Andreas Geiger\textsuperscript{5,10}\quad
  Peng Su\textsuperscript{2$^\dagger$}\quad
  Hongyang Li\textsuperscript{1,4$^\dagger$}\quad
}

\contriblegend{
$^{1}$ HKU \quad
}

\contriblegend{
    \textsuperscript{1} The University of Hong Kong \quad
    \textsuperscript{2} Huawei \quad
    \textsuperscript{3} Shanghai Innovation Institute \quad
    \textsuperscript{4} Archon Robotics  \quad
    \textsuperscript{5} KE:SAI \\
    \textsuperscript{6} NVIDIA Research \quad
    \textsuperscript{7} NTU  \quad
    \textsuperscript{8} valeo.ai \quad
    \textsuperscript{9} Tsinghua University\quad
    \textsuperscript{10} University of Tübingen and Tübingen AI Center \\
    \textsuperscript{*}\,Equal contribution
    \textsuperscript{\dag}\,Corresponding authors \quad
    Contact: \href{}{\texttt{hy@archon.tech}, \texttt{p.su@huawei.com}}
}

\projectlinks{%
  \link{Project Page}{https://opendrivelab.com/WorldEngine/}\quad
  \link{Full Code}{https://github.com/OpenDriveLab/WorldEngine}}

\begin{document}
\maketitle

\begin{abstract}
\setlength{\parindent}{0pt}
Autonomous vehicles must operate safely in the real world, where errors can have severe consequences. 
Although modern end-to-end driving policies excel in routine scenarios, their reliability is limited by the scarcity of safety-critical ``long-tail'' events in real driving datasets. 
These rare interactions define the practical safety boundary of the learned policy, yet they are difficult to collect at scale in the real world.
Here we show that this fundamental limitation can be addressed by post-training pre-trained driving models on synthesized high-stakes interactions.
We introduce World Engine, a generative framework that reconstructs high-fidelity interactive environments from real-world logs and systematically extrapolates them into realistic safety-critical variations.
This paradigm enables reinforcement-based post-training to align policies with safety constraints, circumventing the physical risks inherent in real-world exploration.
On a public benchmark built on nuPlan, World Engine substantially reduces failures in rare safety-critical scenarios and yields significantly larger gains than scaling pre-training data alone.
Furthermore, when deployed on a production-scale autonomous driving system, the resulting policy reduces simulated collisions and demonstrates measurable improvements in on-road testing, showing that post-training on synthesized, safety-critical interactions offers a scalable and effective pathway to safer autonomous driving.
The full codebase suite, including training, is released to the public.
\end{abstract}

\section{Introduction}
Artificial intelligence is undergoing a fundamental transition from digital cognition to Physical AI. Autonomous driving~\cite{hu2023uniad,waymo2025blog,nvidia2025e2escaling} stands as one of the most advanced and socially consequential instances of this shift. Unlike virtual agents, autonomous vehicles perceive, reason, and act directly within the physical world, operating under the strict constraint of irreversibility: errors manifest as physical harm, economic loss, or threats to human safety. As such systems increasingly integrate into daily life, the central challenge is no longer basic task capability, but reliability under safety-critical conditions.

Modern end-to-end autonomous driving systems, trained on millions of kilometres of fleet logs, can now handle the vast majority of everyday scenarios with human-level proficiency~\cite{Kusano2025waymocrash}. However, this average-case performance masks a critical vulnerability: the operational safety boundary is defined not by the common, but by the ``long tail'' of rare events~\cite{liu2024curse}. While uneventful driving is abundant in training data, the abrupt pedestrian crossings, aggressive cut-ins, and complex adversarial interactions that could cause accidents remain statistically sparse.

This scarcity exposes a structural paradox at the heart of autonomous driving: the most safety-critical behaviours must be learned from the scarcest data. Unlike digital domains, where failures can be exhaustively explored, autonomous driving operates under rigid ethical, legal, and social constraints, and society cannot afford to collect safety-critical data at scale. Consequently, the most valuable learning signals residing at the boundary of safe control are systematically missing from naturalistic datasets.

Addressing this paradox remains the central obstacle to safe autonomy. 
Scaling data collection alone yields diminishing returns; accumulating millions of uneventful logs does little to improve robustness in rare moments~\cite{nvidia2025e2escaling,waymo2025motionscaling}. 
Current learning-based systems are thus forced to extrapolate beyond their training distributions, leading to brittle behaviour and unpredictable failures when confronted with novel or compounded risks. 
For autonomous driving to be safely deployable at scale, the field must move beyond passive data accumulation and establish new learning paradigms that explicitly address this long-tail data regime.

A potential solution to this data gap is suggested by the recent evolution of Large Language Models (LLMs).
While scaling pre-training on massive corpora yields broad linguistic competence~\cite{kaplan2020scaling,wei2022emergent}, standard models often struggle with complex, multi-step reasoning---a domain where high-quality natural data is inherently sparse.
The field has responded by moving beyond passive pre-training towards post-training paradigms, which use synthesized reasoning chains by prompting~\cite{wei2022cot} and reinforcement learning~\cite{shao2024deepseekmath} to bridge these gaps. 
This shift is exemplified by systems such as DeepSeek-R1~\cite{guo2025deepseek} and AlphaProof~\cite{hubert2025olympiad}, which achieve superhuman performance in mathematical problem-solving and Olympiad-level competitions. 
Their success validates a crucial principle for autonomous driving: when naturally occurring driving data fails to adequately cover rare but safety-critical situations, learning must be actively guided through synthesis to densify the sparse, high-value regions of the data distribution.

\begin{figure}[!t]
    \centering
    \includegraphics[width=\textwidth]{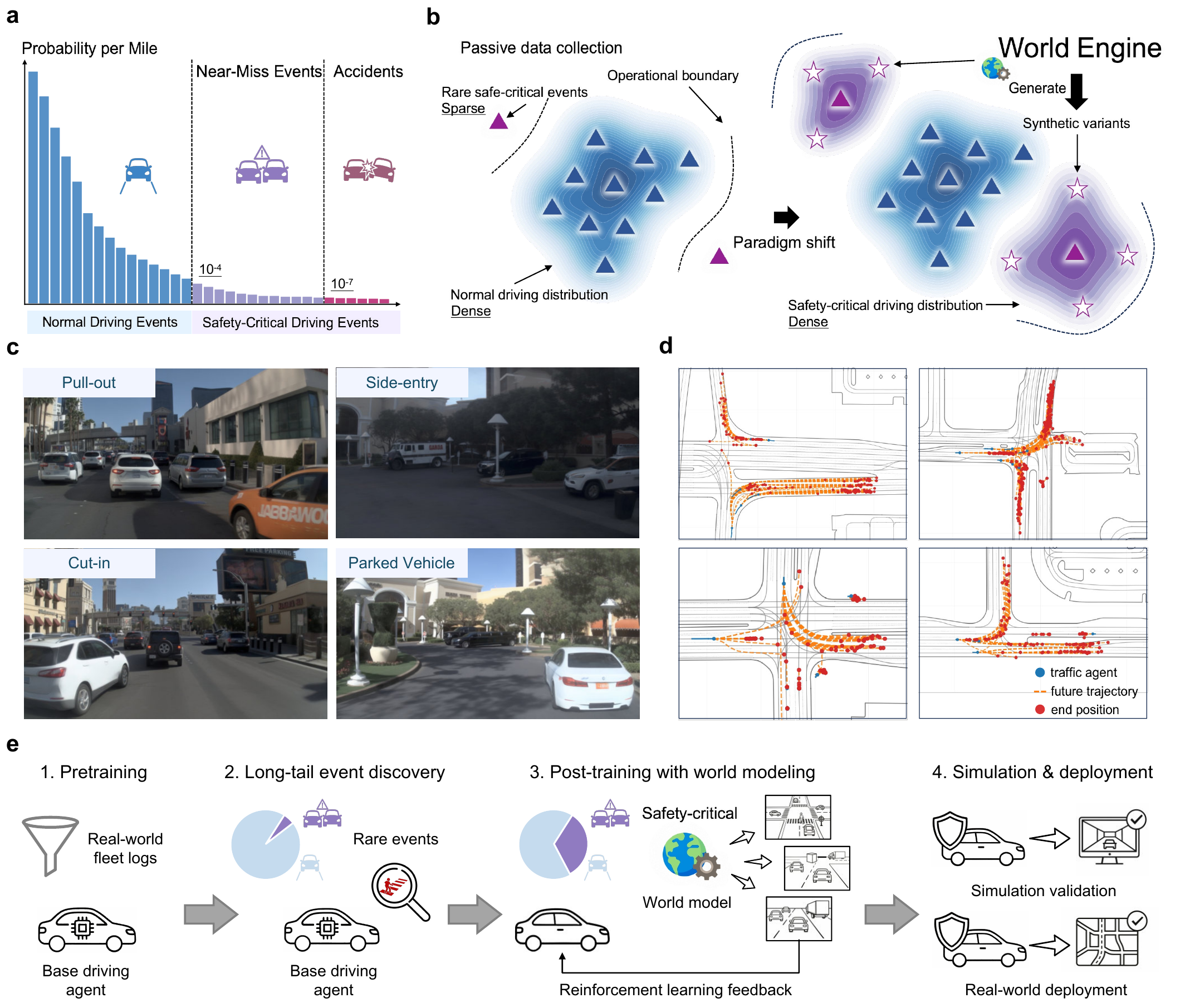}
    \caption{\textbf{World Engine at a Glance.
    }
    \textbf{a,} The curse of scarcity in autonomous driving makes near-miss events and accidents extremely difficult to collect at scale.
    \textbf{b,} Paradigm shift from passive data collection to World Engine. Passive data collection yields dense coverage of common driving scenarios but sparsely samples safety-critical events, leaving them outside the operational boundary of learned models. World Engine instead identifies safety-critical cases and synthesizes diverse variants, converting sparse long-tail events into a dense, learnable safety-critical distribution.
    \textbf{c,} Variants of safety-critical driving scenarios generated by World Engine through photorealistic sensor simulation. 
    \textbf{d,} Behavioural variations synthesized from a single long-tail scenario by the behaviour world model.
    \textbf{e,} Conceptual pipeline of World Engine.
        \textit{1.} Pre-train the agent on real-world driving logs.
        \textit{2.} Identify rare, safety-critical events from the logs.
        \textit{3.} Generate safety-critical variants via world modelling and apply reinforcement learning post-training to improve agent safety.
        \textit{4.} Validate the trained agent in closed-loop simulation and real-world deployment.
    \label{fig:teaser}
    }
\end{figure}

Drawing on this insight, we introduce World Engine (WE), a learning framework that bridges the gap between the rarity of safety-critical events and the need for structured learning in autonomous driving systems.
World Engine discovers failure-prone scenarios from real-world logs, reconstructs them into high-fidelity interactive worlds with diverse traffic variations, and applies reinforcement learning-based post-training to improve planner safety without exposing the real world to additional risk (Fig.~\ref{fig:teaser}).

To demonstrate the effectiveness of World Engine, we apply it to train and evaluate end-to-end autonomous driving agents on a large-scale open-source real-world driving dataset, including sensor data, HD maps and 3D annotations of traffic objects: nuPlan~\cite{nuplan2024}.
We develop a photorealistic driving simulator using state-of-the-art neural rendering techniques to enable closed-loop evaluations of these agents.
On this academic benchmark, we focus on a curated set of safety-critical long-tail scenarios and evaluate models in closed-loop rollouts, where compounding errors and the interactive reactions of other agents can lead to collisions or off-road failures.
Across these rare cases, the full World Engine pipeline substantially improves closed-loop driving quality over the supervised pre-trained baseline, achieving higher success rates under imminent hazards.
Moreover, our data-scaling study reveals that simply increasing pre-training data yields diminishing returns on rare cases, whereas World Engine post-training delivers substantially larger gains than even doubling the pre-training data; extrapolating the scaling trend suggests that it remains competitive even against an order-of-magnitude ($\sim$10$\times$) increase in pre-training data (Fig.~\ref{fig:figure_nuplan_exp}).

Furthermore, we validate the proposed framework on a production-scale autonomous driving system. We train an end-to-end planning model on over 80,000 hours of real-world driving logs, resulting in a substantially strong base model. We then apply reinforcement post-training with World Engine to further improve its performance. We evaluate the model using an industry-grade closed-loop simulation platform with over 10,000 scenarios. 
Results show that, despite the strong baseline, the collision rate is reduced by up to 45.5\% after post-training. We further validate the approach through a 200-kilometre real-world on-road test, achieving zero disengagements and improved safety in safety-critical scenarios.
These results demonstrate that such post-training paradigms can further enhance the safety of already strong autonomous driving systems in real-world settings.

\section{Overview of World Engine}
The autonomous driving problem can be formulated as an end-to-end learning task that maps raw sensor observations to control actions.
In this work, we instantiate this formulation using real-world driving logs (e.g., nuPlan), where sensor data and structured annotations are used to reconstruct interactive environments for training and evaluation.
A key challenge lies in the scarcity of safety-critical and long-tail interactions in real-world driving logs, which fundamentally limits planner robustness (Fig.~\ref{fig:teaser}a).
We introduce World Engine, a unified framework that shifts the closed-loop interactive data distribution toward long-tail scenarios beyond what real-world collection alone can provide, and adapts the end-to-end model through reinforcement learning post-training (Fig.~\ref{fig:teaser}b).
The framework follows a four-stage pipeline: (1) pre-train a base driving agent on large-scale real-world logs and leverage it to discover failure-prone long-tail events, (2) reconstruct each discovered case into a photorealistic interactive simulation via 3D Gaussian Splatting, (3) augment the reconstructed scenarios with diverse traffic variations through a controllable behaviour world model, and (4) refine the agent via reinforcement post-training on the resulting rollouts (Fig.~\ref{fig:teaser}e).
The following subsections describe each stage in detail.

\subsection{Pre-training and Long-tail Event Discovery}
\label{sec:long-tail-discovery}

The first step in World Engine is to identify the scenarios where learning matters most. Rather than synthesizing rare events from scratch or relying on manually designed scenarios, we ground event discovery in real-world driving logs, as they naturally capture complex multi-agent interactions, contextual dependencies, and realistic edge cases that are difficult to faithfully construct through manual design or synthetic perturbations.

We begin by training a base end-to-end autonomous driving model via imitation learning on large-scale driving data. This pre-trained agent serves as both a strong behavioural prior and a diagnostic probe: we feed each logged scenario to the base agent, obtain its planned trajectory, and execute a non-reactive rollout against the logged traffic in a lightweight simulator that operates on 3D bounding boxes and HD maps. Scenarios in which the agent's trajectory collides with logged objects or departs the road are flagged as safety-critical, as they reveal conditions where the current policy fails. These failure-prone cases constitute the long-tail subset that World Engine targets for augmentation.

This discovery strategy offers two advantages. First, the selected events are grounded in real sensor data and real traffic configurations, which ensures that the resulting training distribution remains physically plausible. Second, because the base agent itself defines the boundary of competence, the discovered events are directly aligned with the regions where post-training can yield the largest improvement.

\subsection{Simulation Engine}
\label{sec:simulation-engine}

Once a long-tail event is identified, the next step is to turn it into a simulation environment where the driving agent can practice and learn. We refer to this component as the simulation engine: it reconstructs each discovered scenario into a photorealistic, interactive world and orchestrates closed-loop rollouts where the ego agent and surrounding traffic interact in real time, enabling novel ego trajectories beyond the original recording.

The core of the simulation engine is a 3D Gaussian Splatting (3DGS)-based reconstruction pipeline~\cite{kerbl20233dgs,li2025mtgs}. 
The simulator directly supplies object-level tracks and 3D bounding boxes for all dynamic agents, which are used to decompose the scene into a compositional scene graph separating the static background (roads, buildings, vegetation) from dynamic foreground objects (vehicles, pedestrians, cyclists). Each element is represented by a set of 3D Gaussians fitted to multi-view observations from the driving logs. This decomposition allows independent manipulation of individual objects---repositioning, removing, or altering the trajectory of any traffic participant---while preserving the photorealistic quality of the static surroundings (Fig.~\ref{fig:teaser}c). This capability is critical for generating diverse traffic variations.

A key property of this representation is free-viewpoint rendering: once the scene is reconstructed, it can be rendered from any camera pose, not only those recorded in the original log. This is essential for closed-loop simulation, where both the ego vehicle and surrounding traffic agents may follow trajectories that differ from the original log. As the ego agent takes novel actions and other agents are repositioned or re-planned, the simulation engine produces corresponding sensor observations in real time, maintaining visual fidelity to real-world camera data. The real-time rendering capability enables thousands of rollouts per scenario, which is necessary for reinforcement learning post-training at scale.

\subsection{Behaviour World Model}
\label{sec:behaviour-world-model}

While the simulation engine provides photorealistic sensor observations, the behaviour of surrounding traffic agents must also be modelled to enable meaningful closed-loop interaction. To achieve this, we introduce the behaviour world model, which generates realistic and diverse trajectories for surrounding agents that respond to the ego vehicle's actions.

At its core, the behaviour world model is a learned diffusion model~\cite{zhou2025nexus,li2025omega} that treats multi-agent trajectory generation as a structured denoising process. Given the current scene context---including map topology, historical agent states, and the ego vehicle's planned action---the model generates future trajectories for all surrounding agents simultaneously. The stochastic nature of the diffusion process naturally produces diverse behaviour samples: the same initial condition can yield cooperative, aggressive, or hesitant traffic responses depending on the denoising path.

Beyond stochastic generation, the model supports controllable behaviour synthesis through two mechanisms. First, goal conditioning allows desired endpoints or waypoints to be flexibly specified for individual agents—ranging from explicit constraints for targeted scenario generation to probabilistic sampling for coverage—steering their trajectories toward particular configurations such as cut-in manoeuvres or sudden lane changes.
Second, optimization guidance steers the denoising process at each step by evaluating each candidate trajectory against a desired behavioural objective—for instance, favouring interactions that approach collision thresholds or penalising lane departures—and progressively nudging the generation toward compliant outputs. This requires no retraining of the base model, as the guidance operates solely during sampling.

For flexibility, the simulation framework also supports alternative traffic models: log replay for deterministic reproduction of recorded behaviour, and an Intelligent Driver Model (IDM) \cite{Treiber2000IDM} for rule-based reactive traffic. These modes can be mixed within a single scenario, allowing some agents to follow the learned diffusion model while others replay logged trajectories or follow rule-based control. This combination of stochastic diversity, controllable generation, and flexible traffic modelling enables World Engine to produce counterfactual interactions that probe safety-critical dynamics. A single long-tail scenario can be expanded into hundreds of variations, each presenting different traffic responses that test the ego agent under a broad range of interactive conditions (Fig.~\ref{fig:teaser}d).

\subsection{Reinforcement Post-training}
\label{sec:algorithm-engine}

The simulation engine and behaviour world model together produce diverse closed-loop rollouts from long-tail scenarios. Reinforcement post-training closes the loop by using these rollouts to refine the driving agent.

We formulate the driving task as a partially observable Markov decision process (POMDP), where the agent receives camera observations, maintains internal state estimates, and outputs driving actions. The objective is to learn a policy that maximizes cumulative reward—reflecting safety (e.g., avoiding collisions and maintaining safe distances), comfort (e.g., smooth acceleration with low jerk), and progress (e.g., forward motion along the route)—while remaining close to the pre-trained policy. This balance is achieved through a behaviour-regularized reinforcement learning formulation, where a KL divergence penalty constrains the post-trained policy to stay near the pre-trained prior. The regularization prevents catastrophic forgetting of common driving competence while allowing targeted improvement in safety-critical regimes.

The training data for post-training is drawn from a mixture of two sources: real-world logged trajectories and simulated rollouts generated by World Engine. The real data preserves the distribution of common driving situations, while the simulated data densifies the rare and safety-critical regions. This experience mixing strategy ensures that the agent improves on long-tail events without degrading performance on everyday driving by combining data-level and policy-level regularization: real-world log mixing preserves coverage of common scenarios, while the KL constraint limits deviations from the pre-trained policy.

Within the simulated rollout, hard experience mining further selects the most informative frames for supervision. Not all frames in a rollout are equally valuable: we retain those where the current policy exhibits failure or near-failure behaviour, such as imminent collisions, off-road departures, or large deviations from human driving, and use them as prioritized training samples. By focusing supervision on these hard frames, the agent learns disproportionately from the moments where improvement matters most. A dense reward function provides intermediate feedback across safety, efficiency, and comfort objectives, guiding the reinforcement learning process—together with hard experience mining—toward safe and human-aligned driving behaviour.

The four components described above---pre-training and long-tail event discovery, simulation engine, behaviour world model, and reinforcement post-training---form a closed-loop pipeline. Starting from a pre-trained agent, World Engine discovers its failure modes, reconstructs the corresponding scenarios into interactive worlds, generates diverse traffic variations, and uses the resulting rollouts for reinforcement post-training.

This pipeline enables scalable post-training from rare events grounded in real data. Because every training scenario originates from an actual driving log, the learning signal remains physically plausible. Because the behaviour world model and simulation engine can generate many variations of each scenario, the agent encounters a rich distribution of interactive conditions far beyond what passive data collection can provide. And finally, because reinforcement post-training applies behaviour-regularized reinforcement learning, the resulting policy improves safety in critical regimes without sacrificing competence in common situations.

\section{Methods}
\subsection{Implementation of World Engine}
\label{section:method:worldengine}

In this section, we describe the concrete implementation of World Engine on the publicly available dataset nuPlan, which serves as a reference implementation of our framework and is used for all our controlled ablation studies and quantitative analysis. 
Building upon this foundation, World Engine is also deployed and evaluated on a mass-produced ADAS development stack. 
Although the industrial setup involves additional engineering considerations, it preserves the same conceptual pipeline and learning objectives.
The academic implementation therefore serves as a faithful abstraction of the system used in practice.

\subsection{Simulation Engine}

The photorealistic simulation engine is the key to enabling online exploration and data curation for training an end-to-end planner. It consists of two parts: reconstruction and controllable rendering.

In the reconstruction stage, we reconstruct the driving logs with a 3DGS-based approach. 
Given a set of images and LiDAR points captured within a spatial-temporal driving log, the reconstruction task aims to learn a 3DGS representation that can faithfully reproduce the observed sensor data while recovering the underlying geometric structure. With the 3D bounding box annotation of traffic agents, the dynamic objects are reconstructed separately from static background through a scene graph-based design \cite{Ost2021scenegraph}.
To ensure high-fidelity novel-view rendering, we incorporate dense geometric constraints, including depth and surface normal supervision during reconstruction, ensuring consistent structure and high-quality extrapolated views under novel camera poses.

In the controllable rendering stage, the simulation engine renders the corresponding sensor observations with given input ego pose and sensor extrinsic and intrinsic at timestamp $t$, along with the positions and heading angles of other non-ego vehicles. The 6-degree-of-freedom poses of both ego and non-ego vehicles are calibrated with the ground plane estimated from their trajectories. 
Controllable rendering is achieved by explicitly manipulating the reconstructed dynamic objects in the scene and rendering the camera observations from the updated ego camera location.
Thanks to the efficient rasterization of 3DGS, the simulation engine supports real-time image rendering, enabling closed-loop simulation and scalable online data generation.

We next describe the detailed representation, rendering process, and optimization objectives used in the simulation engine.
\paragraph{Scene representation}
For each driving scene $\tau\sim p$, it is represented as a collection of anisotropic 3D Gaussian primitives $\mathcal{G} = \{ G_i \}_{i=1}^{N},\mathcal{G}\subset \{\mathcal{S},\mathcal{O}\}.$ where each Gaussian $G_i$ is parameterized by its spatial and appearance attributes: $G_i = \{ \mathbf{x}_i, \mathbf{q}_i, \mathbf{s}_i, \alpha_i, \boldsymbol{\beta}_i \}.$ Here, $\mathbf{x}_i \in \mathbb{R}^3$ denotes the centre position, $\mathbf{q}_i \in \mathbb{R}^4$ a unit quaternion of orientation, $\mathbf{s}_i \in \mathbb{R}^3$ the anisotropic scale, $\alpha_i$ the opacity, and $\boldsymbol{\beta}_i$ the spherical harmonic (SH) coefficients of view-dependent colour. Covariance of each Gaussian is constructed as:
$\Sigma_i = R(\mathbf{q}_i)\,\mathrm{diag}(\mathbf{s}_i)^2\,R(\mathbf{q}_i)^\top,$
where $R(\mathbf{q}_i)$ converts the quaternion into a rotation matrix.

To model dynamic driving scenes within a single traversal, Gaussian primitives are organized
into a scene graph that separates static background structure from dynamic scene content.
Specifically, the scene graph consists of two types of nodes:
static nodes $\mathcal{G}_{\text{static}}$, representing stationary elements such as roads and buildings, and
dynamic nodes $\mathcal{G}_{\text{dyn}}$, capturing moving traffic participants that may appear or disappear over time.
This design enables stable reconstruction of the static environment while allowing dynamic objects to be independently manipulated during simulation and rollout.

\paragraph{Appearance modelling} 
We mitigate photometric inconsistency by two-stage calibration. First, LiDAR-guided exposure alignment is applied to correct global brightness variations by matching colours of projected LiDAR points across views. Second, we introduce learnable per-camera affine colour transforms, parameterized by a channel-wise scale and bias, which are shared across time for each camera and optimized jointly with the scene representation. These affine transforms absorb residual camera-specific photometric differences, improving cross-camera consistency during reconstruction.

\paragraph{Rendering and objectives} 
Given camera poses $E_t,K_t$, each 3D Gaussian is projected into image space as a 2D anisotropic Gaussian: $\mathbf{x}_i' = K_t E_t \mathbf{x}_i, \Sigma_i' = J K_t E_t \Sigma_i E_t^\top K_t^\top J^\top,
$ where $J$ denotes the Jacobian of the projection function. Pixel colours are obtained via alpha compositing in depth order:
$\mathbf{c}_p = \sum_i \mathbf{c}_{i,p}\,\alpha_{i,p}
\prod_{j<i} (1 - \alpha_{j,p}),$
where $\mathbf{c}_{i,p}$ is the SH-evaluated colour contribution of Gaussian $G_i$ at pixel $p$, and $\alpha_{i,p}$ its projected opacity. During World Engine rollout, deployed simulation engine generates observation through states $\mathcal{O}=\mathcal{P}_\psi(\mathcal{S})$ in parallel.

The Gaussian parameters are further optimized using a weighted combination of reconstruction and regularization objectives. For reconstruction loss, rendered images are supervised using a combination of $\ell_1$ loss and SSIM. Regularization objectives further introduce depth and normal regularization. Sparse LiDAR depth is incorporated using an inverse-depth loss:
\begin{equation}
\mathcal{L}_{\mathrm{depth}}
=\left|\frac{1}{d_{\mathrm{pred}}}-\frac{1}{d_{\mathrm{LiDAR}}}\right|.
\end{equation}
To improve local geometric consistency and prevent overfitting, a patch-wise normalized cross-correlation loss is applied:
\begin{equation}
\mathcal{L}_{\mathrm{NCC}}
=
1 -\frac{1}{|\Omega|}\sum_{p \in \Omega}\sum_{s=1}^{S^2}\frac{D_{p,s}\,\bar{D}_{p,s}}{\bar{\sigma}_p\,\sigma_p},
\end{equation}
where $\Omega$ denotes the set of depth patches. For normal regularization, rendered normal maps are regularized using pseudo-normals $N$ computed from depth gradients, together with a total variation penalty:
\begin{equation}
\mathcal{L}_{\mathrm{normal}}
=
\left|\hat{N} - N\right|+\mathcal{L}_{\mathrm{TV}}(\hat{N}).
\end{equation}
To prevent degenerate Gaussian shapes, a flattening regularizer is applied:
\begin{equation}
\mathcal{L}_{\mathrm{flatten}}
=\sum_i\max\!\left(\frac{\max(\mathbf{s}_i)}{\mathrm{median}(\mathbf{s}_i)},\, r\right)-r+
\min(\mathbf{s}_i),
\end{equation}
where $\mathbf{s}_i$ denotes the anisotropic scale of Gaussian $G_i$. For transient nodes, Gaussians that fall outside the spatial extent are penalized via:
\begin{equation}
\mathcal{L}_{\mathrm{oob}}=-\frac{1}{\left|\mathcal{G}^{\mathrm{oob}}_{T,k}\right|}
\sum_{G_i \in \mathcal{G}^{\mathrm{oob}}_{T,k}}
\log(1 - \alpha_i).
\end{equation}
The final training loss is the weighted sum of all terms:
$
\mathcal{L}
=
\lambda_r \mathcal{L}_{1}
+
(1 - \lambda_r)\mathcal{L}_{\mathrm{SSIM}}
+
\lambda_{\mathrm{depth}}\mathcal{L}_{\mathrm{depth}}
+
\lambda_{\mathrm{NCC}}\mathcal{L}_{\mathrm{NCC}}
+
\lambda_{\mathrm{normal}}\mathcal{L}_{\mathrm{normal}}
+
\lambda_{\mathrm{flatten}}\mathcal{L}_{\mathrm{flatten}}
+
\lambda_{\mathrm{oob}}\mathcal{L}_{\mathrm{oob}}.
$

\paragraph{Details of reconstruction pipeline}
Reconstruction starts from a driving scenario identified by a key frame within a specific driving log.
Given the key frame timestamp, we extract a spatio-temporal clip consisting of 3 seconds of history and 8 seconds of future frames, using sensor data sampled at 10~Hz.
If the vehicle trajectory within the extracted time window spans less than 50 meters, we extend the end of the clip until either the accumulated trajectory length reaches 50 meters or the end of the driving log is encountered.
This ensures sufficient spatial coverage for stable reconstruction.

To avoid redundant static observations caused by prolonged low-speed or stationary periods, we downsample frames in segments where the ego vehicle speed falls below a predefined threshold. In addition, clips extracted from nearby key frames may partially overlap in time and space. To prevent repeated reconstruction of highly similar content, overlapping clips are merged and reconstructed jointly as a single scene instance.

\paragraph{Handling of image distortion}
We observe that raw camera images exhibit significant lens distortion, which adversely affects both reconstruction quality and pose estimation. All raw images are undistorted using OpenCV \cite{opencv_library} with an optimal undistortion mode that preserves the original field of view during reconstruction. At inference time, we render the image using the optimal undistorted camera intrinsics, and then map it back to the original distorted image space to match the raw image.

\paragraph{Simulation platform and metrics}
We implement a simulation platform that integrates simulation engine with behaviour world model, and serves as the execution backbone for reinforcement post-training and evaluation. The simulator is responsible for generating closed-loop rollouts, computing task-specific metrics and rewards, and recording rollout data for subsequent training and analysis.
To ensure compatibility with diverse end-to-end planners, the simulation platform communicates with external planning models through a lightweight file-based interface. At each simulation step, the simulator serializes sensor observations and scene states to disk, while the planner reads these inputs and returns planned trajectories, which are then executed in the simulator. For traffic agent modelling, the simulator supports either log replay or a reactive Intelligent Driver Model (IDM) \cite{Treiber2000IDM}. Trajectory tracking is executed using an LQR controller with a bicycle dynamic model \cite{Lehtomaki1981LQR,Rajamani2011BicycleModel}. For open-loop evaluation, we use the PDM Score introduced in the NAVSIM benchmark \cite{dauner2024navsim}. For closed-loop evaluation in the World Engine simulator, we report both per-scene success rate (SR) and closed-loop PDM Score (PDMS$^\ast$). SR measures whether the ego vehicle completes the episode without collision or off-road failure, while PDMS$^\ast$ evaluates the quality of the closed-loop trajectory under interactive simulation, providing a complementary progress- and safety-aware metric.

\subsection{Behaviour World Model} 
While the simulation engine provides photorealistic observation of the surrounding environment, the dynamics of traffic agents still need to be modelled to facilitate a comprehensive traffic simulator for end-to-end model training. This behaviour world model goes beyond traditional rule-based or log-replay simulations by utilizing a generative diffusion framework capable of synthesizing both stochastic and adversarial behaviours.

We frame traffic layout simulation as a sequence modelling task, where driving scenarios are represented as structured tokenized states for both agent behaviours and map features, enabling simultaneous prediction of all agent futures. Concretely, the agent behaviours are defined as $\mathbf{x} \in \mathbb{R}^{A\times \mathcal{T} \times D}$, where $A$ denotes the maximum agent capacity, $\mathcal{T}$ represents the physical temporal horizon, and $D$ signifies the dimensionality of agent attributes. Static environmental features are encoded in the map tensor $\mathbf{c}\in \mathbb{R}^{L\times N \times D^{'}}$, representing $L$ lanes with $N$ points per lane and $D^{'}$ attributes (coordinates and types). 

Building upon this vectorized representation, the behavioural world model $T_\theta$ employs a Diffusion Transformer (DiT) that generates the agent tensor $\mathbf{x}$ by reversing a stochastic differential process. Let $\mathbf{x}^0\in \mathcal{X}$ represent a clean agent feature from the distribution $p(\mathbf{x})$. Training begins with an initial state $\mathbf{x}^0$, which undergoes progressive noise injection over time steps $\mathbf{k} = [k_{a,\tau}]\in(0,1]^{A \times \mathcal{T}}$ where each $k_{a, \tau}$ represents the degree of Gaussian noise added to corresponding tokens, until reaching a Gaussian noise distribution at $\mathbf{x}^k$.

The model is optimized by minimizing the mean-square error (MSE):
\begin{align}
\mathbf{x}^\mathbf{k}=\mathbf{\alpha}_\mathbf{k} \mathbf{x}^0+\mathbf{\sigma}_\mathbf{k} \epsilon, \epsilon & \sim \mathcal{N}(\mathbf{0},\mathbf{I}), \mathbf{x}^0 \sim p(\mathbf{x}),
\label{eq:sd_forward}
\\
\forall \mathbf{k}\in(0,1]^{A \times \mathcal{T}},\ \underset{\theta}{\text{min}}\ &\mathbb{E}||(\mathbf{\epsilon}-\epsilon_{\theta}(\mathbf{x}^\mathbf{k};\mathbf{c},\mathbf{k}))\odot\mathbf{m}||_2^2,
\label{eq:sd_train}
\end{align}
where $\mathbf{\alpha}_\mathbf{k}$, $\mathbf{\sigma}_\mathbf{k}$ are scale weights that describe the magnitude of the data $\mathbf{x}^0$ and the noise $\epsilon$ at the denoising step $\mathbf{k}$, $\theta$ parameterizes the denoiser $\epsilon_{\theta}$, and $\mathbf{c}$ is the map tensor guiding the denoising process.
During sampling, all agent tokens are iteratively generated from the standard Gaussian noise with the denoising step.
To enable goal-oriented generation, the model sets a keep mask $\mathbf{m}_{c}$ to ensure targets and past tokens remain fixed during sampling:
\begin{equation}
p(\mathbf{x}^{\mathbf{s}}|\mathbf{x}^{\mathbf{k}}) = \mathcal{N}(\mathbf{x}^{\mathbf{s}}|\mu(\mathbf{x}^\mathbf{k},\mathbf{k}),\Sigma(\mathbf{x}^\mathbf{k},\mathbf{k}))\odot \bar{\mathbf{m}}_{c}+\mathbf{x}^{\mathbf{k}} \odot \mathbf{m}_{c},
\label{eq:label}
\end{equation}
where $\mathbf{s}$ is the next denoising step, $\mu$ and $\Sigma$ are determined by DiT $\epsilon_{\theta}$.

To ensure that generated scenarios align with realistic driving priors, we incorporate classifier guidance~\cite{saharia2022photorealistic}, adjusting the agent tensor $\mathbf{x}$ iteratively to enforce behavioural constraints at each denoising step~\cite{song2020DDIM}. Concretely, we separate overlapping agents along their centreline’s opposite direction to avoid collisions, smooth trajectories, and pull agents toward the nearest lanes for on-road driving.

\subsection{Reinforcement Post-training}

The simulation engine and behaviour world model create diverse counterfactual scenarios, yet these corner cases still need exploratory feedback guided by human priors. The reinforcement post-training stage combines experience curation with reinforcement learning to turn diverse experience into human-aligned improvements.

The reinforcement learning task of autonomous driving can be formulated by the POMDP $\{\mathcal{O},\mathcal{S},\mathcal{A}, r,\gamma\}\subset\mathcal{\tau}$ over future horizon $T$. $o\in\mathcal{O}$ is the observation (i.e., image) from raw sensors. $s\in\mathcal{S}$ denotes the state information of ego vehicle, traffic participants and map. $a\in\mathcal{A}$ is the driving action. $\mathcal{T}_\theta(s_{t+1}|s_t,a_t)$ is the world model; $r: \mathcal{S} \times \mathcal{A} \rightarrow \mathbb{R}$ denotes the shaped reward, and $\pi_\text{ref}$ denotes the pre-trained policy distribution. The optimal driving policy $\pi_\phi(a|o)$ is learned by maximizing the cumulated expected value of reward $R(\tau)=\sum_{t=0}^T\gamma^tr(s_t,a_t)$. We express this objective as a behaviour-regularized reinforcement learning problem:
\begin{equation}
    \max_\phi\mathbb{E}_{\tau\sim p}[R(\tau) - \lambda \sum_{t=0}^T \text{KL}(\pi_\phi(\cdot|o_t)||\pi_\text{ref}(\cdot|o_t))],
\end{equation}
where $\lambda$ is the regularization weight toward the pre-trained driving expert $\pi_\text{ref}$. The experience distribution $p$ is defined as a mixture of real logged trajectories and simulated transitions generated by the world model and simulation engine:
\begin{equation}
    p(\tau)
    = (1-\alpha)\,p_{\mathrm{real}}(\tau)
    + \alpha\,p_{\mathrm{sim}}(\tau \mid \pi_{\phi}, \mathcal{T}_{\theta}, \mathcal{P}_{\psi}).
\end{equation}
Logged transitions $\tau \sim p_{\mathrm{real}}$ are drawn from human driving data. Simulated transitions $p_{\mathrm{sim}}$ are produced by rolling out the policy $a_t \sim \pi_{\phi}(\cdot \mid o_t)$ within the world model $s_{t+1} \sim \mathcal{T}_{\theta}(\cdot \mid s_t, a_t)$  and simulation engine $o_t \sim \mathcal{P}_{\psi}(s_t)$.
This mixture yields a unified experience source and allows the overall objective to be formulated as a regularized policy optimization problem. 

\paragraph{Reward shaping} 
The shaped reward function provides structured feedback that guides the policy toward safe and human-aligned behaviour during post-training. For each trajectory, we compute signals that reflect core driving objectives, including collision avoidance, drivable-area compliance, ego progress along the route, time-to-collision margin, and ride comfort. These rewards help distinguish high-quality behaviours from poor ones and amplify the contribution of informative corner cases. Experiences with higher driving quality naturally yield higher returns, while unsafe or implausible behaviours receive lower values. This ensures the policy learns not only from diverse and counterfactual scenarios but also understands which outcomes are desirable.

\paragraph{Hard experience mining}
Given mixture experience distribution $p$, hard experience mining gathers informative experiences from both logged data and world-model rollouts. We focus on a mixture of scenario clips that reveal rare or safety-critical interactions, such as near collisions, challenging negotiations, recovery manoeuvres, or clear departures from human driving. These mixed samples vary in driving quality, and are assigned different weighted learning curricula. All candidate samples are further checked for physical plausibility. The resulting mixture of high-quality corner cases helps the system learn from instructive experiences while remaining grounded in human driving. 

\subsection{Base End-to-End Driving Model}

In recent years, a variety of end-to-end planning methods~\cite{hu2023uniad,jiang2023vad,chen2024vadv2} have emerged, leveraging onboard sensor data that typically include surround-view cameras and, optionally, LiDAR inputs, together with ego-vehicle information such as IMU poses, localization, and navigation commands.
These models aim to predict the vehicle's future trajectory over several seconds, which is subsequently used by the control module to control vehicle motion. They are generally trained in a supervised manner by imitating human driving behaviour. To further enhance spatial understanding and overall planning performance, these models are often jointly optimized with multiple auxiliary tasks, such as object detection, map element detection, motion prediction, and occupancy estimation.

In this work, we adopt a standardized and robust model architecture. Specifically, we employ a BEVFormer encoder~\cite{li2022bevformer} to process multi-camera inputs and other sensory data, generating a fixed-size BEV feature representation that effectively captures the surrounding spatial context from a bird's-eye-view perspective.
An object tracking decoder and a map segmentation decoder are included for perception supervision following the design of UniAD~\cite{hu2023uniad}. A scoring-based planning decoder~\cite{chen2024vadv2} is included, which selects the best trajectory across a pre-defined trajectory vocabulary, conditioned on BEV feature.

\section{Experiment}
We evaluate the proposed World Engine on two test settings: a visually realistic, challenging closed-loop simulation built upon a real-world dataset, and a large-scale closed-loop simulation and field testing on mass-produced Autonomous Driver Assistance System (ADAS).

\subsection{Implementation Details}
\label{section:method:detail}

\paragraph{NuPlan experiments}
The nuPlan dataset \cite{nuplan2024} comprises 1,282 hours of driving logs, of which approximately 10\% include synchronized sensor data. 
Each vehicle is equipped with 8 surrounding cameras and 5 LiDAR sensors operating at 10 Hz.
We adopt the widely used \textit{navtrain} \cite{dauner2024navsim} split for training, which contains exactly 103,288 scenes, each defined by a key frame with 1.5 seconds of historical context and a 4-second future horizon. Evaluation is performed on the \textit{navtest}, which consists of 12,146 scenes, as well as a rare-event subset of \textit{navtest} comprising 288 failure-prone scenarios. The closed-loop simulation and evaluation is conducted on the rare scenes. 

For data-scaling experiments, the amount of pre-training data is varied from 12.5\% to 100\% of \textit{navtrain}.
Unless otherwise specified, the base agent used for post-training is pre-trained on \textit{navtrain}$_{50pct}$. 
The model backbone consists of a ResNet-50~\cite{he2016resnet} and a feature pyramid network (FPN)~\cite{lin2017FPN} that takes as input 8 camera views with 4 temporal frames per view. 
The extracted features are processed by a six-layer BEVFormer~\cite{li2022bevformer} encoder and a six-layer planning decoder, as well as parallel tracking and mapping decoders~\cite{hu2023uniad}. In total, the model contains 58.3 million trainable parameters.
Pre-training is carried out on eight NVIDIA H100 GPUs and consists of two stages: perception pre-training for 164 hours over 40 epochs, followed by planning pre-training for 15 hours over 8 epochs.
After pre-training, we evaluate the base model on \textit{navtrain}$_{50pct}$ and identify 5,340 long-tail scenes. 
Using World Engine, we generate a dataset of 31,508 frames from these scenarios, which is subsequently used for reinforcement-learning-based post-training. The post-training stage costs 11 hours over 8 epochs.

\paragraph{Huawei ADS experiments}
The pre-training dataset is collected from both internal testing fleets and user vehicles spanning multiple vehicle models.
Each vehicle is equipped with a heterogeneous sensor suite, including ten surrounding cameras, a fused LiDAR, radar, GPS, and an inertial measurement unit (IMU).
After large-scale data mining, automatic labelling, and dataset balancing, the final dataset used for pre-training comprises approximately 80,000 hours of driving data, organized into more than 10 million clips of 25 seconds each.
During post-training, we mix 1.0 million clips generated by World Engine with 5.0 million clips sampled from common driving data.
Pre-training is conducted on Ascend 910B Neural Processing Units (NPUs) for a total of 40,000 NPU-hours, followed by 15,000 NPU-hours of post-training.
For on-road evaluation, the trained policy is deployed on a Huawei AITO M9 vehicle in development mode. In this setting, minimal post-processing is applied: the model outputs are combined with a lightweight control module and used to directly actuate the vehicle chassis.
Although exhaustive route-level deduplication between the pre-training corpus and the on-road test routes is infeasible at this data scale, the specific real-world interactions encountered during testing---including the cut-in event documented in Fig.~\ref{fig:ads_onroad}c,d---are inherently non-reproducible and could not have been present in any logged training data, providing a genuine test of out-of-distribution generalisation.

\begin{figure}[!t]
    \centering
    \includegraphics[width=\textwidth,height=180mm,keepaspectratio]{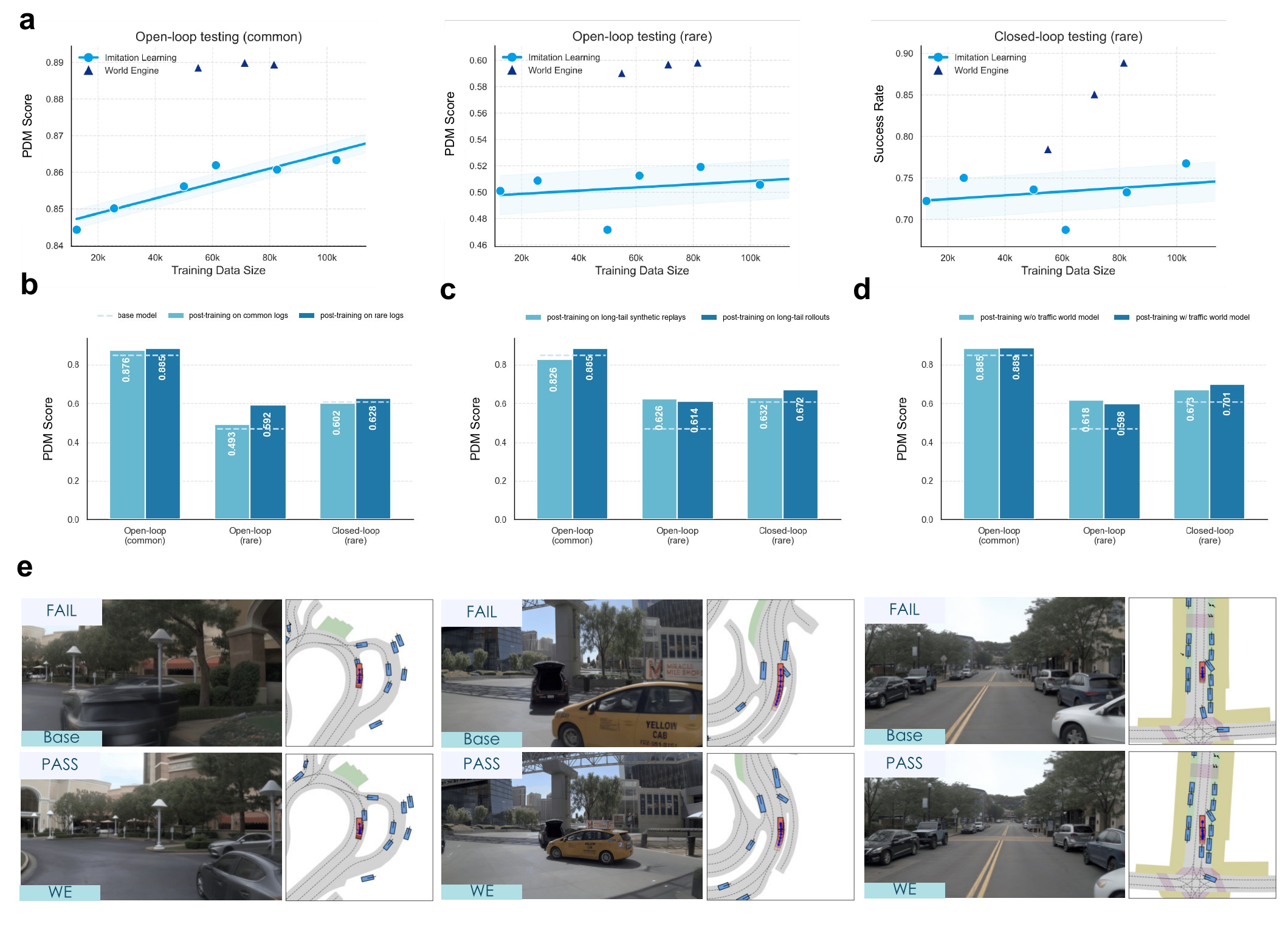}
    \caption{\textbf{Improving autonomous driving systems with World Engine in safety-critical scenarios.}
    \textbf{a,} Effect of scaling the pre-training dataset from 12k to 103k scenes. (Left) Performance on common cases improves predictably with data scale. (Middle, right) Performance gains on rare cases saturate due to the scarcity.
    Starting from a base agent pre-trained on 50k scenes, post-training with safety-oriented rewards leads to substantial open-loop performance improvements on both common and rare cases. Closed-loop performance further improves as the amount of post-training data generated by World Engine increases, achieving comparable gains to those obtained with approximately 14× additional pre-training data.
    \textbf{b,} Post-training on rare-event logs outperforms post-training on common logs on rare open-loop and rare closed-loop PDMS, highlighting the importance of long-tail event discovery.
    \textbf{c,} Post-training on long-tail rollouts better preserves common-case performance and improves rare closed-loop PDMS compared with post-training on long-tail synthetic replays, indicating the value of interactive closed-loop experience beyond replayed behaviours.
    \textbf{d,} Post-training with the behaviour world model further improves rare closed-loop PDMS from 0.673 to 0.701 over post-training without the behaviour world model, showing the benefit of traffic augmentation for closed-loop interaction quality.
    \textbf{e,} Qualitative comparison between the base agent and the agent post-trained with World Engine. For failure cases, we visualize the frame immediately preceding the collision in simulation.
    }
    \label{fig:figure_nuplan_exp}
\end{figure}

\subsection{Safety-critical Closed-loop Simulation}
\begin{figure}[h]
    \centering
    \includegraphics[width=\linewidth]{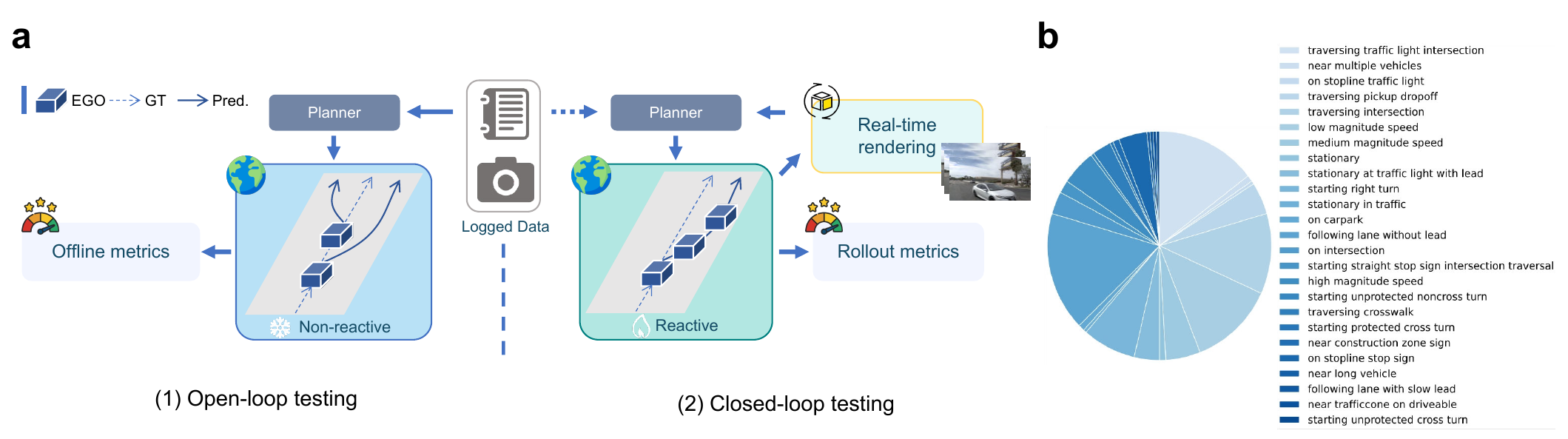}
    \caption{
    \textbf{NuPlan evaluation protocol and safety-critical test-set composition.}
    \textbf{a,} Schematic comparison of open-loop and closed-loop evaluation in nuPlan.
    In open-loop testing, the planner is evaluated on logged observations with offline metrics, without affecting the future evolution of the scene.
    In closed-loop testing, the planner interacts with reactive agents in simulation, and performance is measured using rollout-based metrics under real-time rendering.
    \textbf{b,} Distribution of scenario types in the safety-critical test set used for closed-loop evaluation.
    The set covers a diverse range of challenging traffic situations, including intersections, traffic-light interactions, stop-sign traversals, lane following, and multi-agent encounters.
    }
    \label{fig:extended_figure_nuplan}
\end{figure}

To evaluate the effectiveness of World Engine post-training, we construct a fully open-sourced, reproducible benchmark of safety-critical driving scenes built on the publicly available nuPlan dataset~\cite{nuplan2024}, which comprises 128 hours of real-world driving logs, including sensor data, annotated traffic agents, and high-definition maps. The benchmark is constructed from the official test split of nuPlan, ensuring that all evaluation scenarios are disjoint from the data used for pre-training.
The benchmark focuses on rare cases---short sequences where baseline models are most likely to fail, such as near-collision, off-road deviation, or abrupt interactions with other agents (Fig.~\ref{fig:extended_figure_nuplan}b).
Each test case represents a localized 3D collision-centric scenario extracted from large-scale driving logs, and faithfully reconstructed within the World Engine simulation environment.
Each simulation clip spans 4 seconds, capturing the most failure-prone temporal window in which the driving model must react to an imminent hazard.
We perform open-loop evaluation using the metric PDM Score \cite{dauner2024navsim} and closed-loop evaluation in World Engine simulation using the success rate metric (Fig.~\ref{fig:extended_figure_nuplan}a).
A scenario is considered successful if the vehicle passes the four-second episode without a collision or off-road infraction, reflecting its ability to maintain safe and controllable behaviour under stress.

To examine the data efficiency of World Engine post-training, we design a scaling experiment that varies the amount of pre-training data from 12k to 103k scenes and evaluates each model on three metrics: open-loop PDM Score on common cases, open-loop PDM Score on rare cases, and closed-loop success rate on rare cases (Fig.~\ref{fig:figure_nuplan_exp}a).
We then apply World Engine post-training starting from the 50k-scene base model and compare the resulting gains against the pre-training scaling curve.
While increasing pre-training data yields steady but saturating improvements---particularly on rare cases where safety-critical events are scarce---World Engine post-training produces gains that exceed the scaling trend on all three metrics.
The open-loop improvements confirm that the post-trained agent learns better trajectory planning, while the closed-loop gains further show that this improved planning translates into safer interactive behaviour under compounding dynamics.
In particular, World Engine post-training on the 50k-scene base model already surpasses the performance of a model pre-trained on more than twice as much data. When extrapolating the scaling curve, achieving comparable closed-loop gains through pre-training alone would require approximately an order-of-magnitude more real-world data, highlighting that targeted post-training on synthesized rare events is a far more data-efficient path to safety improvement than scaling passive data collection.

We additionally compare different post-training data sources and interaction models (Fig.~\ref{fig:figure_nuplan_exp}b--d and Table~\ref{tab:method:post_training}). Post-training on common driving logs provides limited or even negative benefit for rare closed-loop evaluation: although it improves common open-loop PDMS, it slightly reduces rare closed-loop PDMS$^\ast$ from 60.98 to 60.20, indicating that common logs alone do not address long-tail interactive failures. Post-training on rare logged scenes improves rare open-loop PDMS from 47.14 to 59.20, but yields only limited closed-loop gains, suggesting that fixed rare logs are insufficient for robust interactive behaviour. Rare synthetic replays further improve rare open-loop PDMS and substantially increase closed-loop success rate, but they yield the lowest ego progress in this comparison, indicating that binary success alone does not fully capture progress-aware closed-loop driving quality. In contrast, rare rollouts without the behaviour world model provide reactive closed-loop experience, better preserve common-case performance, and improve rare closed-loop PDMS$^\ast$ to 67.33. Incorporating augmented traffic interactions through the behaviour world model yields the best overall rare closed-loop performance, achieving the highest PDMS$^\ast$ of 70.12 and the highest success rate of 88.89\%, while maintaining strong common open-loop PDMS. Full quantitative results, including success rate, ego progress, and PDMS$^\ast$, are provided in Table~\ref{tab:method:post_training}.

\begin{table}[t]
\centering
\caption{\textbf{Comparison of post-training paradigms on the nuPlan benchmark.}
We compare different post-training strategies using open-loop PDM Score (PDMS) on common and rare scenarios, and closed-loop metrics on rare scenarios.
SR denotes success rate, EP denotes ego progress, and PDMS$^\ast$ denotes the closed-loop PDM Score.
The base model achieves 85.64 and 47.14 open-loop PDMS on common and rare cases, respectively, with 73.66\% SR, 46.71 EP, and 60.98 PDMS$^\ast$ in rare closed-loop evaluation.
Supervised fine-tuning on rare logs provides only modest improvements over the base model.
Post-training on common logs improves common open-loop PDMS but degrades rare closed-loop SR and PDMS$^\ast$, reducing SR from 73.66\% to 69.62\% and PDMS$^\ast$ from 60.98 to 60.20, highlighting the need for long-tail event discovery.
Post-training on rare logs substantially improves rare open-loop PDMS to 59.20, but does not improve rare closed-loop SR.
Post-training on rare synthetic replays improves rare open-loop PDMS to 62.69 and SR to 87.19\%, but reduces common open-loop PDMS to 82.61 and yields the lowest EP of 32.49, suggesting that binary rare-case success can improve while common-case retention and progress-aware driving quality degrade.
Post-training on rare rollouts without the behaviour world model recovers strong common open-loop performance and achieves the highest EP of 56.74, together with an improved PDMS$^\ast$ of 67.33.
The full World Engine pipeline, which incorporates behaviour-world-model traffic augmentation, achieves the best overall rare closed-loop performance, with the highest SR of 88.89\% and the highest PDMS$^\ast$ of 70.12, while also obtaining the highest common open-loop PDMS of 88.95.
Relative to the base model, World Engine improves rare closed-loop SR by 15.23 percentage points and PDMS$^\ast$ by 9.14; relative to rare rollouts without the behaviour world model, it improves SR by 10.93 percentage points and PDMS$^\ast$ by 2.79.
These results show that combining long-tail event discovery, reactive generated rollouts, and behaviour-model traffic augmentation yields the strongest overall closed-loop safety and driving-quality gains while maintaining strong open-loop performance.}
\label{tab:method:post_training}
\resizebox{\linewidth}{!}{
\begin{tabular}{lccccc}
    \toprule
    Method & \multicolumn{2}{c}{Open-loop PDMS $\uparrow$} & \multicolumn{3}{c}{Closed-loop (rare)} \\
    \cmidrule(lr){2-3} \cmidrule(lr){4-6}
     & common & rare & SR $\uparrow$ & EP $\uparrow$ & PDMS$^\ast$ $\uparrow$ \\
    \midrule
    base model & 85.64 & 47.14 & 73.66 & 46.71 & 60.98 \\
    supervised fine-tuning on rare logs & 87.50 & 52.55 & 74.51 & 47.59 & 61.87 \\
    post-training on common logs & 87.69 & 49.36 & 69.63 & 51.02 & 60.21 \\
    post-training on rare logs & 88.51 & 59.20 & 73.35 & 51.86 & 62.78 \\
    post-training on rare synthetic replays & 82.61 & 62.69 & 87.20 & 32.49 & 63.22 \\
    post-training on rare rollouts w/o Behaviour WM & 88.53 & 61.88 & 77.96 & 56.74 & 67.33 \\
    post-training with World Engine & 88.95 & 59.83 & 88.89 & 47.66 & 70.12 \\
    \bottomrule
\end{tabular}
}
\end{table}

\subsection{Production-scale Driving Validation}

\begin{figure}[!t]
    \centering
    \includegraphics[width=\textwidth,keepaspectratio]{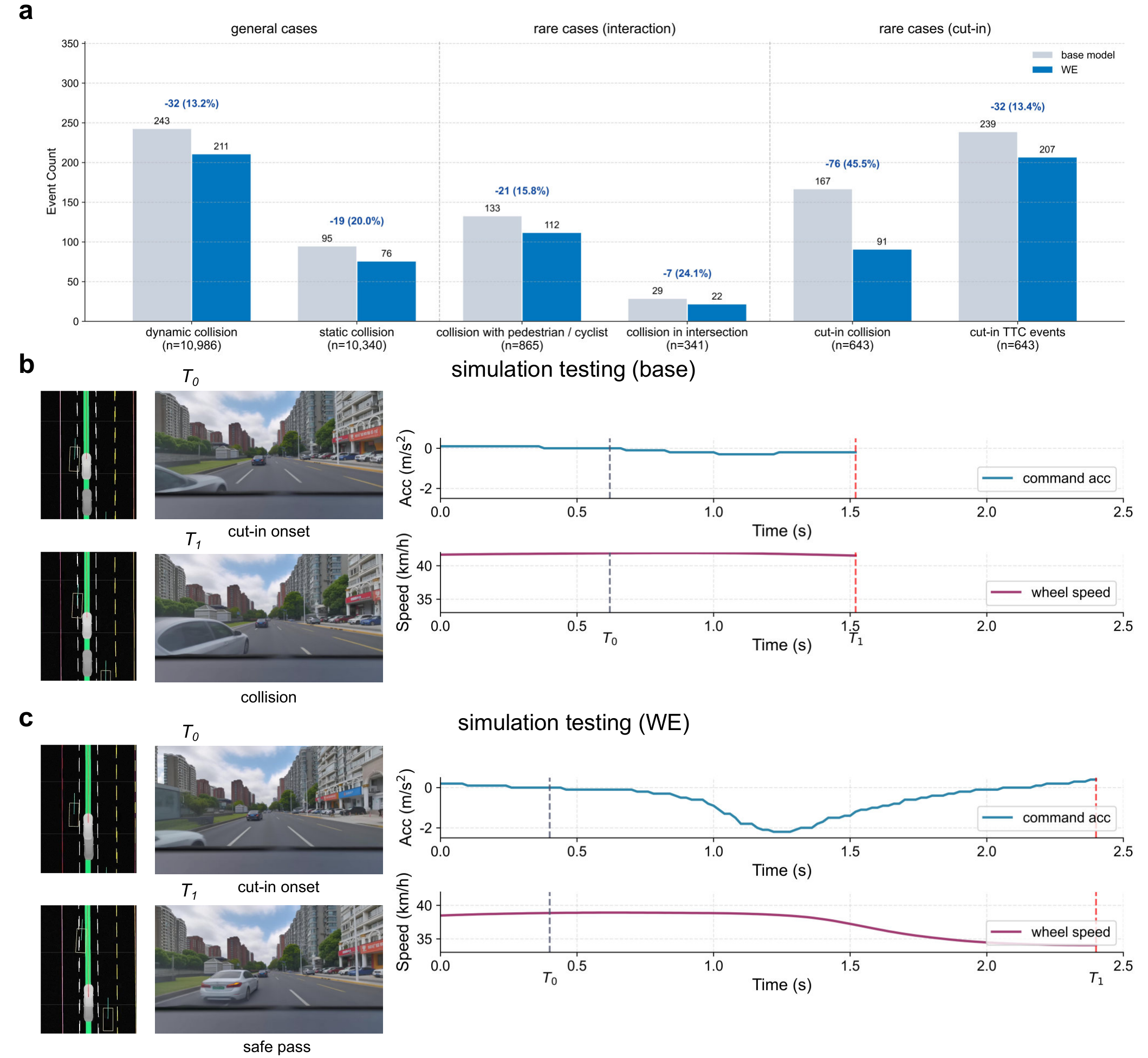}
    \caption{
    \textbf{Testing results in production-level closed-loop simulation.}
    \textbf{a,} World Engine post-training improves key safety metrics over the ADS base model.
    \textbf{b,c,} Representative interaction cases comparing the ADS base and World Engine post-trained models in closed-loop simulation. Each case shows BEV and camera snapshots at key moments ($T_0$ and $T_1$), together with the ego command acceleration and wheel speed.
    }
    \label{fig:ads_sim}
\end{figure}

To validate World Engine at production scale, we apply it to Huawei Advanced Driving System (ADS), an autonomous driving system deployed on over one million vehicles and capable of point-to-point driving in both urban and highway settings.
We leverage the development and testing stack of the system, where an end-to-end model directly takes sensor inputs and produces trajectories to the control module to drive the vehicle.
We first train the base end-to-end model on more than 80,000 hours of real-world driving data collected from over 100 cities.
Following the same pipeline as the earlier experiments, we identify failure-prone scenarios from the training logs, reconstruct them via 3DGS-based rendering, augment traffic interactions through the behaviour world model, and refine the base model through reinforcement post-training.

\begin{figure}[!t]
    \centering
    \includegraphics[width=\textwidth,keepaspectratio]{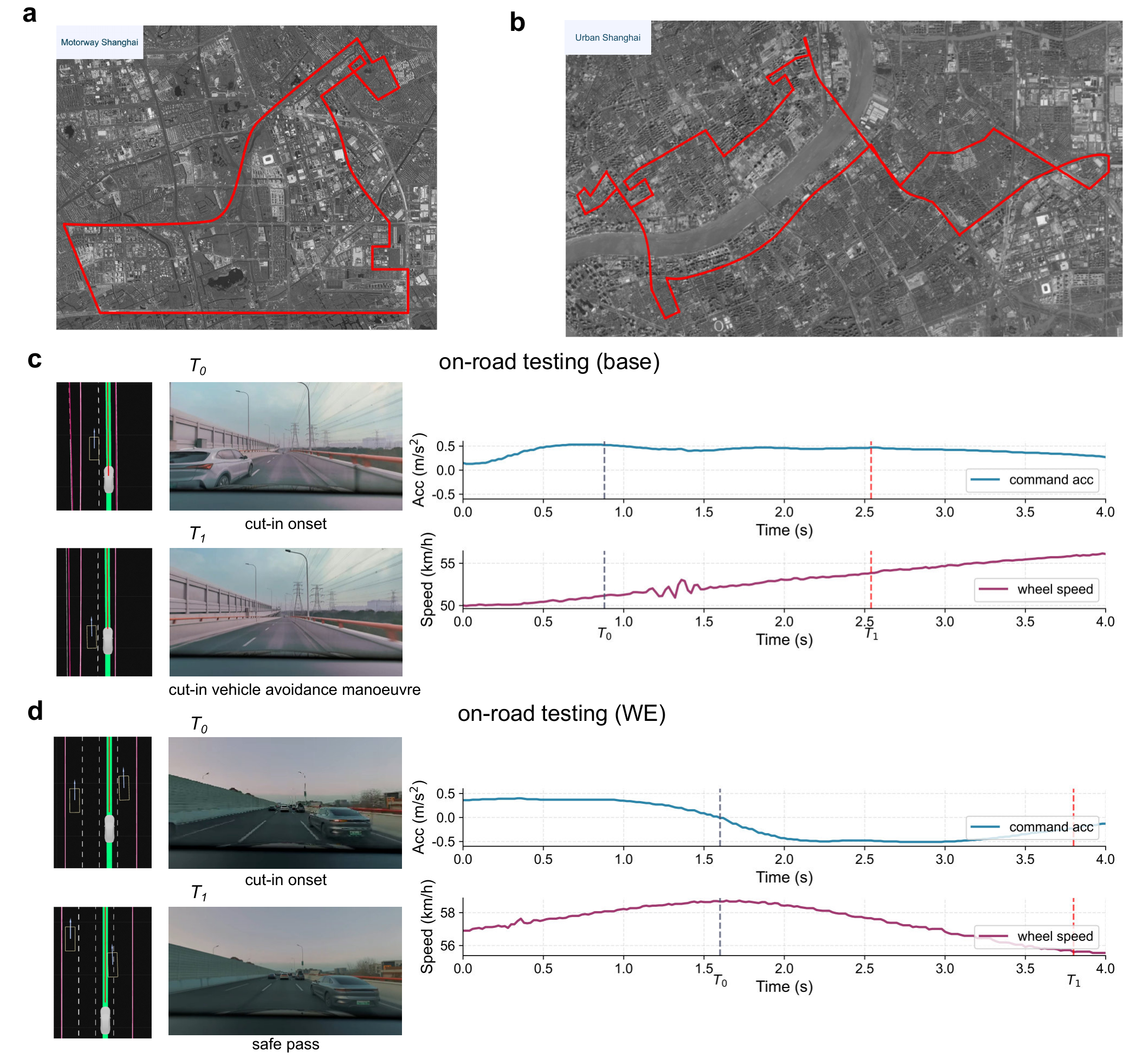}
    \caption{
    \textbf{Scaling World Engine to production self-driving vehicles.}
    \textbf{a,} On-road test route in Shanghai dominated by urban motorways and elevated roads, approximately 65\,km in length, evaluated in two daytime runs.
    \textbf{b,} On-road test route in urban Shanghai dominated by town and residential roads, approximately 70\,km in length, evaluated in one nighttime run.
    \textbf{c,d,} Representative interaction cases comparing the ADS base and World Engine post-trained models in on-road tests.
Each case shows BEV and camera snapshots at key moments ($T_0$ and $T_1$), together with the ego command acceleration and wheel speed.
    }
    \label{fig:ads_onroad}
\end{figure}

We evaluate the post-trained model using the industrial quality assurance system, which performs asynchronous, hardware-in-the-loop closed-loop simulations on production on-board hardware.
In each simulation, rendered sensor streams are fed to the end-to-end model running on the vehicle's computing unit, forming a full sensor-to-control closed loop.
Each test scenario runs for approximately 20 seconds; across all scenarios the simulation totals over 60 hours, equivalent to roughly 3,000\,km of driving---all consisting of eventful, interaction-rich situations rather than uneventful cruising.
Across six safety metrics spanning general and rare driving scenarios, World Engine post-training consistently reduces failure events (Fig.~\ref{fig:ads_sim}a).
In rare interaction scenarios (1,206 test cases), collisions with pedestrians and cyclists decrease by 15.8\% and collisions at intersections decrease by 24.1\%.
In rare cut-in scenarios (643 test cases), cut-in collisions decrease by 45.5\% and time-to-collision events decrease by 13.4\%.
These safety gains do not come at the cost of general driving performance: across 10,986 common cases, dynamic collisions decrease by 13.2\% and static collisions by 20.0\%, indicating that long-tail post-training also improves everyday driving competence.
A representative simulation case is shown in Fig.~\ref{fig:ads_sim}b,c: the base model fails to brake in time and collides with a cut-in vehicle, while the post-trained model decelerates proactively and avoids contact.

\begin{figure*}[t!]
    \centering
    \includegraphics[width=\textwidth,keepaspectratio]{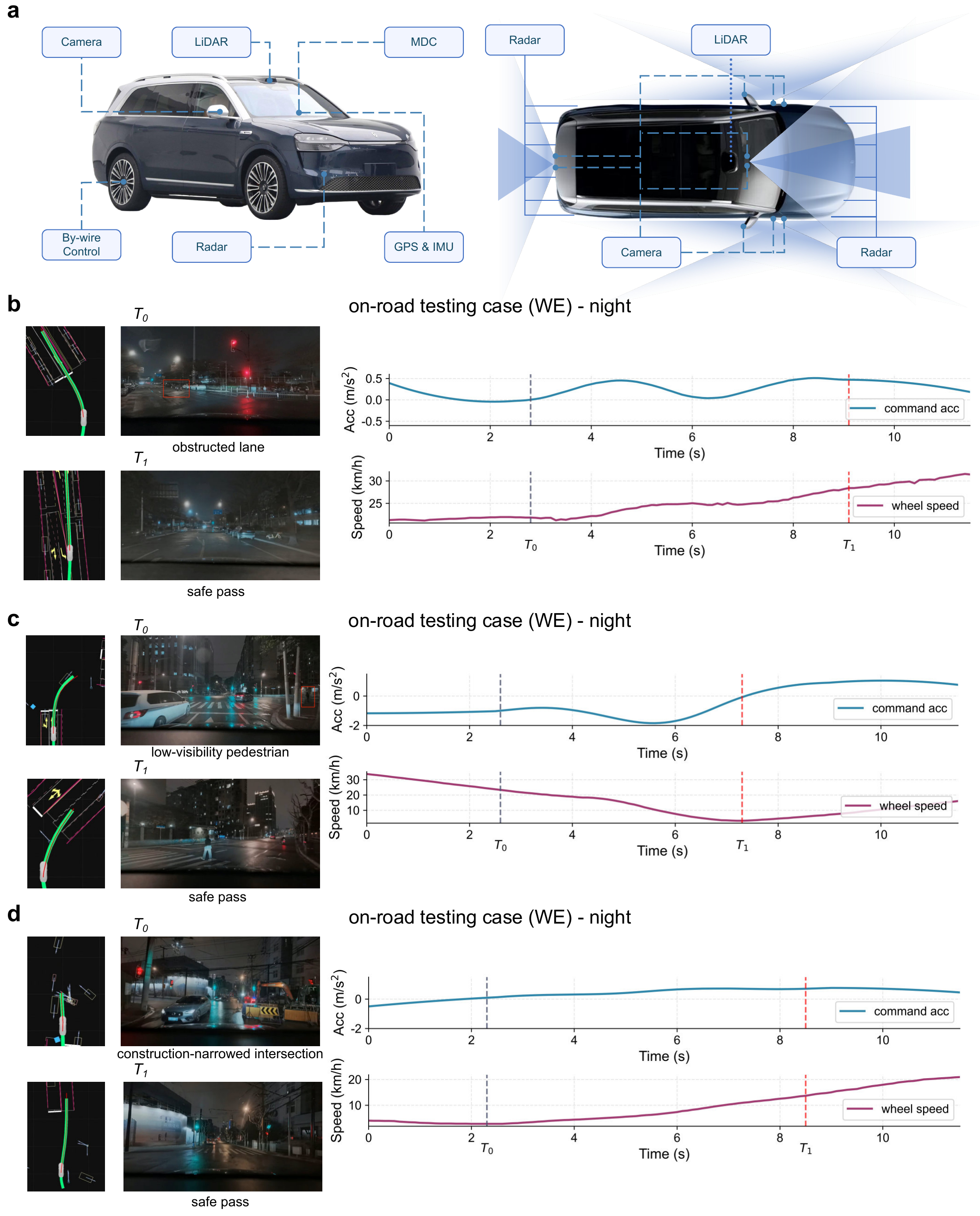}
    \caption{
    \textbf{Scaling World Engine to nighttime driving scenarios.}
    \textbf{a,} Sensor configuration of the test vehicle.
    \textbf{b,} Nighttime obstructed-lane scenario, showing safe passage around the obstruction.
    \textbf{c,} Nighttime low-visibility pedestrian crossing, showing safe yielding and passage.
    \textbf{d,} Nighttime construction-narrowed intersection, showing safe negotiation through the constrained roadway.
Each case shows BEV and camera snapshots at key moments ($T_0$ and $T_1$), together with the ego command acceleration and wheel speed.}
    \label{fig:extend_ads_onroad}
\end{figure*}

We further conduct on-road testing over multiple runs totalling approximately 200\,km across routes in Shanghai (Fig.~\ref{fig:ads_onroad}a,b).
Across all three runs, the post-trained model completes the full 200\,km with zero disengagements, whereas the base model triggers one safety-critical intervention.
Even a single intervention over 200\,km remains a substantial reliability failure for deployment, particularly because it occurs in a rare but safety-critical cut-in interaction.
Fig.~\ref{fig:ads_onroad}c,d illustrate this incident: an adjacent vehicle, unaware of the test vehicle approaching from behind, begins to merge into the ego lane.
The base model fails to respond to the cut-in and even attempts to accelerate; the adjacent vehicle is forced to abort its lane change abruptly to avoid a collision (Fig.~\ref{fig:ads_onroad}c)---a situation that poses clear safety risks.
Under a similar condition, the post-trained model recognizes the cut-in early and smoothly adjusts its speed, allowing the ego vehicle to pass safely without requiring evasive action from either party (Fig.~\ref{fig:ads_onroad}d).
These results, together with additional nighttime driving scenarios (Fig.~\ref{fig:extend_ads_onroad}), demonstrate that the safety improvements from World Engine transfer from simulation to real-world deployment on production vehicles.

\section{Conclusion and Discussion}
World Engine reframes autonomous driving learning under the long-tail regime as a post-training problem: instead of relying on passive fleet-scale collection to eventually observe rare events, it discovers failure-prone scenarios from real logs, reconstructs them into interactive photorealistic worlds, expands them through counterfactual traffic variations, and improves end-to-end planners via reinforcement learning.
Across a safety-critical closed-loop benchmark reconstructed from nuPlan, post-training in World Engine improves success rate and trajectory feasibility, and can deliver closed-loop gains comparable to adding substantially more pre-training data, highlighting a data-efficient path for safety alignment when rare events are the bottleneck.
Scaling to a mass-produced ADAS stack, World Engine reduces failure rates in a safety-critical benchmark (up to 45.5\% in cut-in tests) while maintaining performance on common cases, suggesting that targeted long-tail post-training can improve safety without sacrificing everyday driving competence.
From a practical standpoint, these results carry significant implications for the industry. Our data-scaling analysis shows that World Engine post-training can match the safety gains of approximately 10$\times$ more pre-training data when extrapolating the scaling curve, which in practice would require proportionally larger fleets, longer collection campaigns, and substantially higher annotation costs.
By shifting the focus from passive data accumulation to targeted synthesis and reinforcement on rare events, World Engine offers a more cost-effective path to resolving long-tail safety problems and can substantially shorten the iteration cycle for improving autonomous driving systems.

\paragraph{Limitations.}
We note several limitations of the current framework.
First, the long-tail event discovery stage can only identify failure modes that are already present in the collected driving logs. If a safety-critical scenario type has never been recorded---for example, an unusual road geometry or an entirely novel agent behaviour---it cannot be discovered or reconstructed by the current pipeline. Extending discovery beyond logged data, for instance through procedural scenario generation or adversarial search in a learned latent space, remains an open direction.
Second, the fidelity of World Engine is bounded by its two simulation components. The 3DGS-based simulation engine produces high-quality images near the recorded trajectories but degrades when the simulated ego path deviates substantially from the original log, introducing visual artefacts that may affect policy learning. The behaviour world model, while capable of diverse traffic generation, does not yet capture all real-world interaction patterns with high fidelity, particularly those involving pedestrians, cyclists, or unstructured road users. Narrowing this sim-to-real gap is essential for further improving the transfer of post-training gains to deployment.
Third, the reward function used in reinforcement post-training is based on principled but manually defined signals, such as collision avoidance, lane adherence, and route progress. These rewards encode general safety and driving objectives, yet they may not capture all the nuances of human driving preferences. Developing learned or verifiable reward functions that can adapt to more complex driving norms is a promising avenue for future work.

\paragraph{Scalability and iterative refinement.}
The current framework is validated with a single round of post-training. A natural extension is iterative refinement, where the improved agent is re-evaluated to discover new failure modes, and successive rounds of World Engine post-training are applied. However, in our experiments with the relatively small model (58.3M parameters), we observe that multiple rounds of post-training tend to destabilize the policy, likely because the model lacks sufficient capacity to absorb corrections without interfering with previously learned behaviours. Whether larger models or more sophisticated regularization strategies can support stable multi-round post-training remains an open question.
A related observation concerns how failure modes evolve as the base model scales. In the Huawei ADS experiments, where the base model is trained on over 80,000 hours of data, the proportion of scenarios identified as long-tail events is substantially smaller than in the academic setting. Importantly, even with fewer discovered long-tail events, post-training still yields meaningful safety improvements. This finding indicates that World Engine can continue to provide gains as base models grow stronger, targeting an increasingly narrow but consequential set of failure modes.

\paragraph{Future work on world modelling.}
In this work, we develop a world modelling framework for autonomous driving that combines 3DGS-based neural rendering with a behaviour world model to generate realistic and controllable behaviours of surrounding agents. 
The proposed approach satisfies key requirements for post-training, including controllability, realism, efficiency, and robustness. 
A current limitation arises from imperfections in 3DGS reconstruction, particularly when large deviations occur between simulated rollout trajectories and the original real-world trajectories. 
As video generation technology continues to advance, this world modelling framework could be extended to video-based world models, which have shown promise in embodied navigation and beyond-the-view scene imagination~\cite{zhang2026sparse}, offering improved realism, controllability, and diversity.

\paragraph{Towards general Physical AI.}
The core insight behind World Engine---that safety-critical learning requires active synthesis rather than passive collection---is not specific to driving. Any Physical AI system that must operate reliably in the real world faces the same structural challenge: the most consequential failures are the hardest to observe in natural data. 
Robot manipulation systems, legged locomotion policies, and surgical robots all share this property. 
In these areas, the long tail of rare but high-consequence events defines the practical safety boundary, and is systematically under-represented in training data.
The discover--world-modelling--post-train pipeline introduced in this work offers a general template for addressing this challenge. 
Given a pre-trained policy and a set of recorded task executions, one can identify failure-prone episodes, reconstruct them into interactive worlds, generate diverse variations, and apply reinforcement post-training to improve robustness in the identified regimes. 
The specific instantiation of each stage will vary across domains---3DGS rendering and behaviour world models may be replaced by video world models~\cite{nvidia2025cosmos,yang2026rise} or physics-based simulation~\cite{dosovitskiy2017carla,jia2024bench,mu2025robotwin} depending on the application---but the overall logic remains the same: ground learning in real failures, expand coverage through controllable synthesis, and refine the policy through targeted reinforcement.
We believe that this paradigm of post-training on synthesized rare events, bridging pre-training on broad natural distributions with targeted reinforcement on safety-critical regimes, represents a promising direction for building reliably safe Physical AI systems.

\section{Acknowledgments}
This work is in part supported by the JC STEM Lab of Autonomous Intelligent Systems funded by The Hong Kong Jockey Club Charities Trust. 
We thank Chonghao Sima, Huijie Wang, Jiazhi Yang, Haochen Tian, Yihang Qiu from OpenDriveLab for comments and discussions.
We thank Jianzhang Yang, Hengyu Lu and colleagues at Huawei Inc. for their assistance with experiments on the Huawei ADS.

\clearpage
\bibliographystyle{unsrtnat}
\bibliography{bibliography_short,references}

\clearpage
\bookmarksetup{startatroot}
\phantomsection
\pdfbookmark[1]{Supplementary Materials}{supplementary-materials}
{\noindent\Large\itshape Supplementary Materials\par}

\setcounter{section}{0}
\setcounter{subsection}{0}
\setcounter{figure}{0}
\setcounter{table}{0}

\renewcommand{\thesection}{\Alph{section}}
\renewcommand{\thesubsection}{\Alph{section}.\arabic{subsection}}
\renewcommand{\thefigure}{S\arabic{figure}}
\renewcommand{\thetable}{S\arabic{table}}

\renewcommand{\theHsection}{supp.\Alph{section}}
\renewcommand{\theHsubsection}{supp.\Alph{section}.\arabic{subsection}}
\renewcommand{\theHfigure}{suppfig.\arabic{figure}}
\renewcommand{\theHtable}{supptab.\arabic{table}}

\input{./X_supp}

\end{document}

%% file: X_supp.tex
\section{Related work}
\label{sec:supp:related}

\subsection{Data Engine for Autonomous Systems}
Data Engine describes the closed-loop data collection pipeline~\cite{mitchell2018never}, where models continuously improve by driving the selection, generation and curation of new data~\cite{van2022three}. In autonomous driving, this idea is already visible in industrial ``data flywheels'' that couple large fleets with targeted mining and re-annotation~\cite{li2023open}. Data-centric engines trigger from mispredictions in production, operating on focused data collection and relabelling before retraining and redeployment~\cite{abbaspour2025dataset,scaleai_data_engine_2023,tesla_aiday_2022}. Similar patterns are designed that identify model weaknesses, prioritize difficult scenes for annotation and close the loop between evaluation and dataset growth~\cite{lai2024uncertainty,appliedintuition_sds_automotive}. Specifically for perception, auto-labelling frameworks~\cite{liang2024aide} formalize open vocabulary annotations~\cite{kirillov2023segment}. VLMs~\cite{li2023blip}, incremental learning~\cite{thengane2022clip}, and model in-the-loop techniques~\cite{qi2021offboard} are integrated to query, label and retrain on the most informative samples to steadily increase coverage and label quality over time. The paralleled line of work focuses on simulation and rendering for system verification~\cite{muhammad2020deep}. Earlier works start from manually designed physical engine-based platforms~\cite{dosovitskiy2017carla}. Behaviour agents simulate traffic participants with varying degrees of realism~\cite{montali2023waymo} or safety criticality~\cite{feng2023dense,feng2021intelligent}. Reconstructed driving scenes from real-world clips are used to generate multi-sensor observations for closed-loop validation~\cite{gao2024vista,yang2025resim,li2025uniscene}, and several platforms offer controllable scenario generation~\cite{yang2023unisim,ding2024realgen,gao2025magicdrive}. Recent efforts also incorporate regulation- or standards-based testing, including NCAP procedures, NHTSA typologies~\cite{jia2024bench2drive}, and traffic-rule compliance~\cite{yu2024online}, to ensure legitimacy and safety coverage. Beyond driving, robotics employs analogous data flywheels that use HIL learning~\cite{luo2025precise}, residual tuning~\cite{chi2024iterative}, and self-filtered instruction generation to iteratively expand and enhance experience~\cite{gupta2021embodied}. Conceptually, the emerging view is that agents should treat the world as an open-ended training set rather than a fixed corpus~\cite{silver2025welcome}, and that explicit data engines are essential within autonomous driving stacks to address the remaining coverage gaps at scale.

\subsection{Learning Pipeline in Autonomous Driving}

Autonomous driving systems have traditionally been developed as modular stacks that decouple perception, prediction and planning, with each component trained independently~\cite{badue2021self}. Several works attempt to integrate prediction and planning under open or closed-loop formulations~\cite{hagedorn2024integration}. Still, compounding errors that propagate across modules remain a persistent challenge. End-to-end systems address this by learning a direct mapping from raw sensory inputs to future actions through unified learning frameworks~\cite{chen2023e2esurvey}. Textual and agent-based priors further support interpretability and reasoning through vision-language models~\cite{sima2024drivelm,tian2024drivevlm,hwang2024emma,yang2024llmdrive} or vision-language-action modelling~\cite{li2025drivevla,chen2024driving,Yang_2026_CVPR,zhou2025autovla,jia2026guidedvla}.
Learning typically begins in a pre-training setting using open-loop imitation learning on large-scale human demonstrations~\cite{pan2020imitation}. While deployed under closed-loop real-world driving, open-loop pre-trained systems suffer from covariate shift and causal confusions~\cite{cai2022closing}. To alleviate this mismatch, prior work explores limited forms of end-to-end closed-loop pre-training, such as DAgger rollout co-training~\cite{ross2011dagger,laskey2017dart,tian2026simscale}, on-demand expert~\cite{peng2023learning,wu2023humanintheloop}, or RLVR-style feedback scoring~\cite{li2025recogdrive,guo2025ipad,zou2025diffusiondrivev2}. Still, they remain confined in training stability, temporal scope, and interaction diversity~\cite{karkus2025beyondbc}.
This motivates a growing body of work on post-training~\cite{guo2025deepseek}.
A pre-trained policy is further refined through interaction-aware objectives~\cite{lu2023imitation}. Existing post-training techniques span reinforcement learning fine-tuning in simulation~\cite{peng2024improving,zhang2025carplanner}, residual or adapter-based policy updates~\cite{wang2025adawm}, distillation from planner-based or privileged experts~\cite{zhang2024asymmetricselfplay}, test-time adaptation~\cite{wang2024drivingfuture,li2025end}, and model-based reinforcement learning leveraging learned world models~\cite{yang2025rawdrive,jiang2025irlvla,li2024think2drive}. By operating beyond static demonstrations, post-training enables the policy to correct compounding errors, reason over long-horizon consequences, and improve performance in rare, safety-critical, or counterfactual scenarios that are poorly represented in human data~\cite{gao2025rad,yan2026ad,ni2025recondreamerrl}. Nevertheless, current post-training approaches often rely on full-policy optimization, dense reward design, or extensive closed-loop interaction, which can introduce instability, degrade pre-trained behaviours, or overfit to specific simulators or environments. Heavy reliance on online interaction further limits scalability, while unconstrained policy updates risk forgetting the human-aligned priors established during pre-training.

\subsection{Comparisons with Related Work}
Data engines, simulation, and post-training have become established paradigms for extending learning beyond static demonstrations. In most existing systems, however, these components play a supporting role. Data engines are primarily used for dataset expansion or system validation, improving perception and coverage but operating largely in open-loop settings that do not capture the causal effects of policy actions. Simulation and neural rendering enable controllable replay or perturbation of real-world scenes, providing valuable tools for verification and stress testing, yet generated scenarios are typically used to assess robustness rather than to directly guide learning, and behaviour generation is rarely coupled to the evolving weaknesses of the policy. Post-training methods further refine pre-trained planners through reinforcement learning, residual adaptation, or model-based optimization, but are often constrained by simulator bias, dense reward design, or unstable full-policy updates, and are commonly applied as narrow fine-tuning steps tied to specific environments or benchmarks.

In contrast, our approach is designed to be both active and integrated, addressing limitations that arise when data generation, simulation, and learning are treated as separate stages. World Engine actively generates experience by rolling out the policy in photorealistic, interactive environments, exposing it to the causal consequences of its own decisions rather than relying on passively collected data or static scenario perturbations. By combining neural scene reconstruction with behaviour-level world modelling, the system synthesizes diverse and counterfactual interactions that systematically populate the long-tail regimes where natural driving data are sparse. Crucially, this experience is not consumed indiscriminately. Experience generation, curation, and optimization are tightly coupled within a single closed loop, allowing reinforcement post-training to prioritize informative corner cases and reinforce behaviour using human-aligned objectives. This integration ensures that post-training is both targeted and stable, avoiding the brittleness associated with narrow fine-tuning or unconstrained reinforcement learning. Rather than serving as an auxiliary refinement or verification tool, post-training in our framework becomes a scalable mechanism for reshaping end-to-end driving behaviour, enabling the system to improve robustness and safety in rare, interaction-driven scenarios while remaining anchored to human intent.



\section{Supplementary methods}
\label{sec:supp:worldengine}

\subsection{Behaviour World Model}
The behaviour world model enhances the generative diffusion model by incorporating individual noise levels to generate the realistic, reactive and controllable traffic behaviour, which underpins the World Engine system.



\subsubsection{Scene Vectorized Representation}

Driving scenarios are encoded as structured token representations, consisting of an \textit{Agent Tensor} for dynamic entities and a \textit{Map Tensor} for static environment features. We denote the agent tensor as $\mathbf{x} \in \mathbb{R}^{A\times \mathcal{T} \times D}$, where $A$ is the maximum number of agents, $\mathcal{T}$ is the number of physical time steps, and $D$ is the dimension of agent attributes. The attribute for each agent includes positional coordinates ($x$, $y$), heading ($sin_{\alpha}$, $cos_{\alpha}$), velocities ($v_x$, $v_y$), and dimensions ($l$, $w$). A valid mask $\mathbf{m} \in \mathbb{B}^{A\times \mathcal{T}}$ is initialized to indicate which agents in the agent tensor $\mathbf{x}$ are valid at each time step. As for the map information, the map tensor $\mathbf{c}\in \mathbb{R}^{L\times N \times D^{'}}$ is used to represent the lanes' conditions, where $L$, $N$, and $D^{'}$ stand for the number of lanes, points per lane, and attributes (coordinates and types), respectively. Based on the vectorized representation, sequential modelling of driving scenes can be expressed as generating the future scene tensor $\mathbf{x}\odot \mathbf{m}_{:,\tau:,:}$ given the current time step $\tau<\mathcal{T}$, historical scene tensor $\mathbf{x}_{:,:\tau,:}$, and global map tensor $\mathbf{c}$. To simplify the model's learning task, all feature channels are normalized with corresponding means and deviations before concatenating.

\subsubsection{Decoupled Noise Modelling}
We introduce decoupled triaxial mask modelling, where independent noise levels align across agent indices, temporal time steps, and denoising steps, thereby resolving the causal complexity and efficiency bottlenecks in modelling reactive traffic with diffusion model.
Specifically, we denote $\boldsymbol{x}_{a,\tau}^{k_{a,\tau}}$ as the token of $a$-th agent $\boldsymbol{x}_{a,\tau}$ within $\mathbf{x}^{k_{a,\tau}}$ at noise level $k_{a,\tau}$ under the forward diffusion process; $\boldsymbol{x}_{a,\tau}^0$ and $\boldsymbol{x}_{a,\tau}^T$ represent the clean 
token and the pure noise.
The noise level matrix $\mathbf{k} = [k_{a,\tau}]\in(0,1]^{A \times \mathcal{T}}$ of the sequence is assigned a random matrix, representing the degrees of Gaussian noise $\mathbf{\epsilon}$ added to corresponding tokens.
%
%
The optimizing process of the scene generation model can be written as:
\begin{equation}
    \forall ~\mathbf{k} \in (0,1]^{A\times \mathcal{T}},\ \underset{\theta}{\text{min}}\ \mathbb{E}||(\mathbf{\epsilon}-\epsilon_{\theta}(g(\mathbf{x}^0,\mathbf{k});\mathbf{c},\mathbf{k}))||_2^2,
\end{equation}
where $g$ represents the function that adds noise $\mathbf{\epsilon}$ to $\mathbf{x}^0$ using matrix $\mathbf{k}$, where each token is masked to varying degrees.
The model is learned by completing the full sequence from soft-masked tokens, following information from low-noise tokens when generating other parts.
During sampling, setting history and goals to low noise and others to high noise ensures conditional guidance in scene generation.
\subsubsection{Classifier Guidance for Human-behaviour Alignment}
Diffusion models can generate unrealistic driving scenarios due to randomness, requiring human-guided constraints to enhance scene quality. Specifically, we consider three human-behaviour rubrics:

1) Collision avoidance: At each step $t$, if two vehicles' bounding boxes overlap, they are pushed apart along their centre-connecting line.
It can be written as the following equation:
\begin{equation}
   \begin{aligned}
f_{\text{collision}}(\mathbf{x}^t, t) &= \big[ \mathbf{x}^{t}_{\text{loc}}, \mathbf{x}^{t, 3:d} \big],\\
\text{where}\ 
\mathbf{x}^{t}_{\text{loc}} &\gets \mathbf{x}^{t}_{\text{loc}} + \lambda_t \sum_{i \neq j} \mathbb{I} \{B(\mathbf{x}^t_i) \cap B(\mathbf{x}^t_j) \neq \varnothing \} \cdot 
\frac{\mathbf{x}^{t}_{i,\text{loc}} - \mathbf{x}^{t}_{\text{j,loc}}}{\|\mathbf{x}^{t}_{i,\text{loc}} - \mathbf{x}^{t}_{j,\text{loc}}\|},
\end{aligned} 
\end{equation}

where $\lambda_t$ is a scalar coefficient used to control the extent of separation at time $t$.
$\mathbb{I}$ is an indicator function that takes the value 1 when the bounding boxes of vehicle $i$ and vehicle $j$ overlap and 0 otherwise.
$B$ is the function used to form the vehicle's bounding box.
The fractional term represents the unit direction vector of the centreline between vehicle $i$ and vehicle $j$.

2) Comfort: Enforcing smooth longitudinal and lateral accelerations by averaging adjacent trajectory points.
\begin{equation}
\begin{aligned}
f_{\text{comfort}}(\mathbf{x}^t, t) &= \big[ \mathbf{x}^{t}_{\text{loc}}, \mathbf{x}^{t, 3:d} \big],\\
\text{where}\ 
\mathbf{x}^{t}_{\text{loc}} &\gets \mathbf{x}^{t}_{\text{loc}} - \lambda_t \mathbf{a}^{t}, \\
\mathbf{a}^{t} &= \frac{1}{2} (\mathbf{x}^{t}_{\tau-1,\text{loc}} - 2\mathbf{x}^{t}_{\tau, \text{loc}} + \mathbf{x}^{t}_{\tau+1, \text{loc}}).
\end{aligned}
\end{equation}
First, the longitudinal and lateral accelerations $a^{t}$ are approximated using the second-order difference at time $\tau$ and smoothed by averaging adjacent trajectory points.  
Then, the trajectory is refined by subtracting a proportion $\lambda_t$ of the acceleration, reducing abrupt speed changes for smoother motion.

3) On-road driving: Pull the vehicle toward the nearest centreline point when it strays too far.
\begin{equation}
\begin{aligned}
f_{\text{on road}}(\mathbf{x}^t, t) &= \big[ \mathbf{x}^{t}_{\text{loc}}, \mathbf{x}^{t, 3:d} \big],\\
\text{where}\
\mathbf{x}_{i,\text{loc}}^{t} & \gets \mathbf{x}_{i,\text{loc}}^{t} + \lambda_t \mathbb{I} \{\|\mathbf{x}_{i,\text{loc}}^{t} - \mathbf{c}_i^t\| > d_{\text{th}}\} \cdot (\mathbf{c}_i^t - \mathbf{x}_{i, \text{loc}}^{t}), \\
\mathbf{c}_i^t &= \operatorname{argmin}_{l, n} \|\mathbf{x}_{i, \text{loc}}^{t} - \mathbf{c}_{l, n, \text{loc}}\|.
\end{aligned}
\end{equation}

The vehicle identifies the closest lane point $\mathbf{c}_{i}^t$ among all points $\mathbf{c} \in \mathbb{R}^{L \times N \times D}$ by minimizing the Euclidean distance using $ \arg\min_{l,n}$.  
When the deviation exceeds the threshold $ d_{\text{th}} $, the vehicle adjusts its position by moving from $ \mathbf{x}^t_{i,\text{loc}} $ toward the closest centreline point $ \mathbf{c}_i^t $, with the adjustment magnitude controlled by $\lambda_t$.

\subsubsection{Inference Mode} 
The core objective of the inference mode is to leverage the world model's robust goal-following capabilities to synthesize a diverse set of safety-critical driving scenes. By doing so, we augment our dataset with valuable corner cases that are often missing from standard data. We employ two distinct generation strategies to achieve this: \textit{Scenario Copy} and \textit{Intent Attack}.

\textbf{Scenario copy strategy.}
This strategy aims to replicate existing critical scenarios while introducing controlled variations to prevent exact duplication. The process begins by identifying all relevant agents---including the ego vehicle---within a predefined distance threshold (e.g., 10 meters). For each identified agent, we perturb the original trajectory endpoint by adding a random displacement vector. Specifically, the new goal point is sampled uniformly from a circular region with a radius of 1 meter centred on the original goal, formulated as:
\begin{equation}
G_{new} = G_{original} + \delta, \quad \text{where } \lVert \delta \rVert < 1m.
\end{equation}
The traffic world model is then conditioned on these perturbed goals to generate complete, interactive trajectories for all agents. By applying this controlled perturbation, we produce new scenarios that maintain the fundamental interaction patterns of the original data while introducing meaningful behavioural diversity.

\textbf{Intent attack strategy.}
Complementing the copy strategy, Intent attack is designed to proactively generate safety-critical scenarios by manipulating the intentions of surrounding agents. The primary motivation is to address the scarcity of high-risk interactions—such as aggressive cut-ins—by creating adversarial situations that stress-test the ego vehicle's planner.

In this approach, we randomly select one ``adversarial agent'' from the $K$-nearest neighbours to the ego vehicle (e.g., $K=3$). The goal point of this agent is forcibly reassigned to a location near the ego vehicle’s original goal, formulated as:
\begin{equation}
G_{\text{adversary}} = G_{\text{ego}} + \epsilon, \quad \text{where } |\epsilon| \leq \Delta.
\end{equation}
Here, $\Delta$ defines a small proximity boundary (e.g., 1--2 meters) to ensure a high likelihood of spatial conflict. The behaviour world model then generates a full scene rollout conditioned on this adversarial goal. This results in a realistic scenario where the selected agent intentionally interferes with the ego vehicle’s path, effectively augmenting the dataset with intent-ambiguous edge cases.

\textbf{Validation and post-processing.}
To ensure physical plausibility and behavioural validity, every generated trajectory undergoes rigorous post-processing. Since raw model outputs may occasionally violate kinematic constraints, we first refine trajectories using an LQR-based tracker. The optimized trajectories are then subjected to two critical checks: 
\textbf{(1) Collision detection:} Verifying that no unavoidable collisions occur between agents in the initial state. 
\textbf{(2) Drivable area verification:} Ensuring all vehicles remain within legal road boundaries.
Only scenarios that pass both validation checks are deemed successful and added to the dataset.

\textbf{Iterative generation framework.}
These strategies operate within an iterative framework to systematically construct a diverse repository of scenarios. During each iteration, a strategy is randomly selected and applied to a seed scenario. The process terminates when either: \textbf{(1)} the number of validated scenarios reaches a target $N$ (e.g., $N = 10$), or \textbf{(2)} consecutive generation failures exceed a failure threshold $F$ (e.g., $F = 30$). This dual-termination mechanism ensures computational efficiency while preventing excessive resource expenditure on difficult seeds, achieving an optimal balance between dataset scale, diversity, and quality.

\subsection{Reinforcement Post-training}


To enable continuous and active improvement of driving performance, reinforcement post-training is built within a scalable framework. The framework integrates curated corner-case sampling and verifiable reward-shaping modules under well-defined optimization objectives, allowing efficient and scalable post-training through large-scale rollouts.

\subsubsection{Data Sampler} Although large-scale fleet and counterfactual data provide abundant driving experience, the resulting distribution is highly imbalanced. To address this, we adopt a structured corner-case curation and curriculum weighting strategy that explicitly controls both \emph{what} experiences are sampled and \emph{how} strongly they influence policy updates during post-training.

Corner-case trajectories are curated through a structured data pipeline that integrates heterogeneous sources and long-horizon driving semantics. Candidate experiences are collected and sampled from different sources $u\in\mathcal{U}$, such as real-world fleet logs, user data, and user-triggered events (e.g., disengagements or near-miss alerts), offline mining of logged data using risk- and uncertainty-based detectors, and counterfactual rollouts generated by the simulation engine and behaviour world model. Each trajectory $\tau$ is associated with a semantic category $c \in \mathcal{C}$ by a series of semantics such as interaction, driving style, or driving manoeuvres. To balance coverage and safety emphasis, sampling is performed from a hierarchical mixture over sources and categories.
\begin{equation}
p_{\mathrm{cur}}(\tau)=\sum_{u\in\mathcal{U}}\sum_{c\in\mathcal{C}}
p_\mathcal{U}(u)\,p_{\mathcal{C}}(c\mid u)\,p(\tau\mid c,u),
\end{equation}
where $p_{\mathcal{U}}$ and $p_{\mathcal{C}}$ encode internal sampling weights that maintain diversity in sources and styles. Within each category, planning manoeuvres vary in instructional value. We therefore assign a trajectory-level quality score curricula:
\begin{equation}
w(\tau)=\mathrm{clip}\!\left(
\frac{q(\tau)-q_{\min}}{q_{\max}-q_{\min}},
\,w_{\min},\,w_{\max}
\right),
\end{equation}
which limits the influence of rare but noisy corner cases. Curriculum weighting is then incorporated directly into policy optimization by reweighting the objectives. This formulation cleanly separates coverage control through source- and category-level sampling from trajectory-level curriculum weighting, enabling scalable and conservative post-training focused on informative corner cases.

\paragraph{Verifiable reward modules}
To ensure that policy improvement remains aligned with human driving principles during large-scale post-training, we design the reward function $R(\tau)$ as a set of verifiable modules with explicit structure and bounded influence. Rather than relying on a single scalar reward, we decompose feedback into interpretable components and combine them through weighted aggregation and multiplicative gating, allowing safety-critical constraints to dominate optimization while softer objectives shape preferences among valid behaviours:
\begin{equation}
r(s_t,a_t)=\prod_{m \in \mathcal{M}} r_m(s_t,a_t)\times\frac{1}{\sum_{w \in \mathcal{W}}
\beta_w}\sum_{w \in \mathcal{W}}
\beta_w \, r_w(s_t,a_t).
\end{equation}

\paragraph{Safety-critical-constraint gates ($\mathcal{M}$).}
We use two binary or partially penalized gates to enforce non-negotiable safety requirements (Table~\ref{tab:hard_reward_terms}).

\begin{table}[h]
\centering
\caption{Safety-critical-constraint gates used in reinforcement post-training.}
\label{tab:hard_reward_terms}
\begin{tabular}{p{3.6cm} p{8.6cm}}
\toprule
Component & Definition \\
\midrule
No-at-fault collision ($r_{\mathrm{col}}$) & Returns 1.0 if the ego vehicle incurs no at-fault collision; 0.5 for a single collision with a static obstacle (for example, a parked vehicle or road furniture); and 0.0 for any collision with a dynamic agent.  \\
Drivable-area compliance ($r_{\mathrm{dac}}$) & Returns 1.0 if all corners of the ego bounding box remain within legal drivable regions (lanes, intersections, or parking areas) throughout the step, and 0.0 otherwise. \\
\bottomrule
\end{tabular}
\end{table}

Because these safety-critical-constraint gates multiplicatively modulate the remaining reward terms, safety-critical violations sharply suppress the overall reward regardless of driving quality elsewhere in the rollout.

\paragraph{Soft driving objectives ($\mathcal{W}$).}
In addition to the hard safety gates, three active components contribute to the weighted average (Table~\ref{tab:soft_reward_terms}).

\begin{table}[h]
\centering
\caption{Soft driving objectives used in reinforcement post-training.}
\label{tab:soft_reward_terms}
\begin{tabular}{p{3.2cm} c p{8.8cm}}
\toprule
Component & Weight $\beta_w$ & Definition \\
\midrule
Ego progress ($r_{\mathrm{prog}}$) & 5.0 & Distance advanced along the route centreline (m), normalized by an estimated safe upper bound computed using the search-based planner PDM-Closed \cite{Dauner2023pdm}, and clipped to $[0,1]$. \\
Time-to-collision ($r_{\mathrm{ttc}}$) & 5.0 & Binary safety margin: 1.0 if the predicted time-to-collision with all surrounding agents exceeds 1.0\,s under constant-velocity extrapolation, and 0.0 otherwise. \\
Comfort ($r_{\mathrm{comf}}$) & 2.0 & Binary ride-quality check: 1.0 if longitudinal acceleration, lateral acceleration, and steering-angle rate all remain within human-calibrated comfort bounds, and 0.0 otherwise. \\
\bottomrule
\end{tabular}
\end{table}

Ego progress and time-to-collision are assigned equal and highest weight to jointly optimize forward task completion and proactive collision avoidance, two objectives that can otherwise trade off against each other in aggressive cut-in or intersection scenarios. Comfort receives a lower weight to shape policy preference among safe behaviours without penalizing necessary evasive manoeuvres. Empirically, equal weighting between progress and time-to-collision helped avoid degenerate solutions in which the agent either halts indefinitely to minimize collision risk or accelerates through near-miss events to maximize progress.


\subsubsection{Rollout Platform}

The platform supports safe and scalable deployment of post-trained agents while continuously harvesting large volumes of experience. Leveraging the World Engine, massive rollouts are generated via high-throughput rendering and behaviour world modelling, enabling efficient collection of diverse and rare interaction data. 
Safety signals and performance metrics are closely monitored, and any detected degradation triggers immediate rollback. Deployment is progressively expanded, enabling continuous fleet data collection, engine rollout, curation, and policy improvement loop through World Engine, while maintaining safety and reliability at scale.

\subsection{Simulation Engine: Closed-loop Platform}

\begin{figure}[t]
    \centering
    \includegraphics[width=\textwidth]{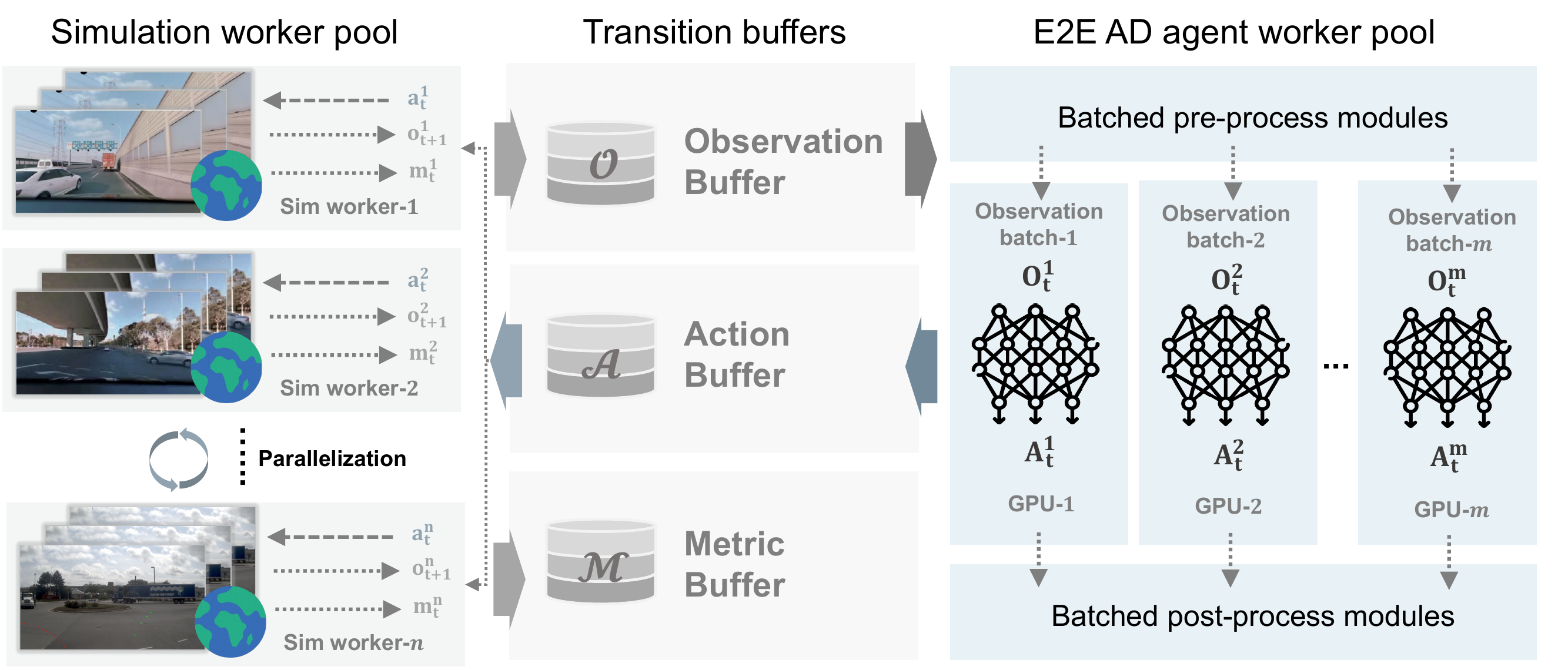}
    \caption{\textbf{Overall pipeline of closed-loop autonomous driving testing platform.} }
    \label{fig:supp:sim_platform}
\end{figure}

To enable scalable and interactive evaluation of end-to-end driving agents over logged scenarios, we introduce a closed-loop autonomous driving testing platform. Unlike existing simulator-centric E2E evaluation frameworks and neural-rendering-based simulators, our platform targets large-scale driving logs and supports richer interaction patterns. By leveraging multi-traversal neural rendering, the platform enables massively parallel rollouts while preserving high-fidelity perceptual feedback in regions critical to decision making. Interaction fidelity is configurable, ranging from fully interactive traffic to optional non-reactive or weakly reactive IDM agents, allowing controlled and reproducible evaluation. 
This facilitates systematic assessment of end-to-end policies over long horizons, capturing safety, robustness, and compounding-error effects, while aligned with real-world driving data.

\subsubsection{Platform Setup}
Following the closed-loop formulations of each simulation step, the platform receives actions $a_t$ from the end-to-end policy $\pi_\theta$, and provides high-fidelity observation $o_{t+1}$ and states $s_{t+1}$ for next step rollout. Per-step evaluation metrics ${m}_t$ and termination identifier $\mathbf{d}_t$ are also buffered within each simulation loop. Once terminated, testing metrics are calculated.

\paragraph{State and observation}
Each evaluation scenario is initialized from logged driving data, maintaining a set of privileged states $s_t$ that encode the ego vehicle and all traffic participants over $T_\text{sim}$ simulation steps, together with road topology and dynamic traffic signals. Rendered scene representations are pre-deployed and cached, enabling efficient synthesis of observations $\mathcal{O}$ under the ego vehicle’s extrinsic parameters for each simulated state.

\paragraph{Controller} 
The end-to-end policy $\pi_\theta$ operates at a fixed control frequency and outputs either the planned waypoints or direct control commands of acceleration $a$ and steering angle $\delta$. For planning waypoints, the platform offers an LQR controller that track the waypoint into executable control.

\paragraph{State simulation}
For ego vehicle, given controlling command $[a_t,\delta_t]$ through controller, the platform propagates the ego state with a discrete-time kinematic bicycle model:
\begin{equation}
\begin{aligned}
x_{t+1} &= x_t + v_t \cos(\psi_t)\,\Delta t, \\
y_{t+1} &= y_t + v_t \sin(\psi_t)\,\Delta t, \\
\psi_{t+1} &= \psi_t + \frac{v_t}{L}\tan(\delta_t)\,\Delta t, \\
v_{t+1} &= \mathrm{clip}\!\left(v_t + a_t \Delta t,\; 0,\; v_{\max}\right), \\
\delta_{t+1} &= \mathrm{clip}\!\left(\delta_t + \dot{\delta}_t \Delta t,\; -\delta_{\max},\; \delta_{\max}\right),
\end{aligned}
\end{equation}
where $L$ is the wheelbase, and $\mathrm{clip}(\cdot)$ enforces physical bounds on speed and steering. For non-ego traffic participants, we support multiple simulation modes depending on the evaluation setting.
Specifically, traffic agents can (i) follow replayed states from logged trajectories for non-reactive simulation, (ii) be controlled by an Intelligent Driver Model (IDM)~\cite{Treiber2000IDM} to enable reactive longitudinal behaviour along each agent's local lane direction, or (iii) be generated by a learned behaviour world model with diverse traffic interaction. In closed-loop evaluation, the traffic agents are controlled by IDM.

\paragraph{Termination detection}
Termination conditions $\mathbf{d}_t$ are evaluated online at each step using state-based safety monitors. An episode terminates if any hard constraint is violated, including collision, deviation from navigated routes, or rule violation. Episodes also terminate upon successful route completion or upon reaching a predefined time horizon. 

\paragraph{Platform communication}
To enable scalable closed-loop evaluation, the platform is implemented as a distributed, modular system that decouples policy inference from state simulation and rendering. Simulation workers advance scenario states and synthesize observations, while inference workers asynchronously consume batched observations from buffered simulation pool, and return corresponding actions. Each evaluation component is registered through a lightweight builder that exposes standardized hooks at key execution points.

\subsubsection{Testing Metrics}

We evaluate planning performance under both open-loop and closed-loop settings, using metrics tailored to the characteristics of each evaluation protocol.

\paragraph{Open-loop metrics}
For open-loop evaluation, we follow NAVSIM \cite{dauner2024navsim} and adopt the \emph{Predictive Driver Model Score (PDMS)} as the primary metric. PDMS is designed to provide a comprehensive assessment of driving quality under a short-horizon, non-reactive simulation, where background agents strictly follow their recorded future trajectories and the ego vehicle is executed via an LQR controller over a 4-second horizon at 10~Hz.

PDMS aggregates multiple complementary sub-metrics that capture safety, legality, efficiency, and comfort:

\begin{itemize}
    \item \textbf{No-at-fault Collisions (NC).}  
    This metric penalizes collisions for which the ego vehicle is deemed responsible. At-fault cases include (i) collisions with stationary objects, (ii) ego-front collisions with any traffic agent, and (iii) ego-side collisions occurring in intersections or multiple lanes. The score is discrete-valued: $\mathrm{NC}=1$ if no at-fault collision occurs, $0.5$ for a single collision with a static class, and $0$ otherwise.

    \item \textbf{Drivable Area Compliance (DAC).}  
    DAC enforces adherence to traffic rules by requiring the ego vehicle to remain within drivable regions (e.g., lanes, intersections, or parking areas). If any corner of the ego bounding box leaves the drivable area, the score is set to $\mathrm{DAC}=0$; otherwise, $\mathrm{DAC}=1$.

    \item \textbf{Time-to-Collision (TTC).}  
    TTC measures near-miss risk by estimating the minimum predicted time before collision under a constant-velocity extrapolation. In NAVSIM, the ego vehicle is projected forward with a fixed velocity and heading at 0.3 s steps, and collisions with surrounding agents are checked. The score is binary: $\mathrm{TTC}=1$ if the minimum TTC exceeds 0.9~s, and $0$ otherwise.

    \item \textbf{Ego Progress (EP).}  
    EP evaluates how effectively the ego vehicle advances along the intended route. Progress is normalized by an estimated safe upper bound, obtained via a search-based planner (PDM-Closed \cite{Dauner2023pdm}). The resulting ratio is clipped to $[0,1]$, with negligible or negative progress discarded.

    \item \textbf{Comfort (C).}  
    The comfort metric verifies that kinematic quantities such as acceleration and jerk remain within human-calibrated thresholds, following the nuPlan framework. If all thresholds are satisfied, $\mathrm{C}=1$; otherwise, $\mathrm{C}=0$.
\end{itemize}

The final PDMS is computed as:
\begin{equation}
\mathrm{PDMS}
= \mathrm{NC} \cdot \mathrm{DAC}
\cdot \frac{5\,\mathrm{TTC} + 5\,\mathrm{EP} + 2\,\mathrm{C}}{12}.
\end{equation}

\paragraph{Closed-loop metrics}
While PDMS is well-suited for open-loop evaluation, closed-loop execution introduces compounding interactions over time that are not fully captured by a single-step predictive score. Therefore, we adopt two episode-level metrics for closed-loop evaluation.

\begin{itemize}
    \item \textbf{Success Rate.}  
    For each closed-loop rollout $\tau$, we define a binary validity indicator that requires the ego vehicle to complete the rollout without triggering either a collision or a drivable-area violation:
    \begin{equation}
    \mathrm{Success}(\tau)
    = \prod_{t=0}^{T}
    \mathbb{I}(\mathrm{NC}_t)\,
    \mathbb{I}(\mathrm{DAC}_t).
    \end{equation}
    The reported Success Rate is the dataset average, directly reflecting safety-critical feasibility in closed-loop control.

    \item \textbf{PDMS$^\ast$.}  
    To bridge open-loop scoring with closed-loop behaviour, we introduce \emph{PDMS$^\ast$}.
    We treat the posterior closed-loop trajectories---including both the ego vehicle and surrounding traffic agents generated during the rollout---as if they were the planner's predicted trajectories at $t=0$, and re-evaluate them using the PDMS formulation.
    PDMS$^\ast$ differs from standard PDMS in two key aspects.
    First, while standard PDMS normalizes Ego Progress (EP) against an upper bound derived from the PDM-Closed planner, PDMS$^\ast$ instead uses the human-driven ground-truth trajectory as the reference.
    Second, unlike standard PDMS which assumes non-reactive background agents, PDMS$^\ast$ evaluates the ego trajectory under reactive traffic behaviours that emerge during closed-loop interaction.
    These modifications enable a more faithful assessment of efficiency and safety in closed-loop settings, while preserving the structure and interpretability of PDMS.

\end{itemize}
\subsubsection{Computational Cost}

We provide a high-level estimate of the runtime cost of the closed-loop platform based on the representative experiment executed on 8 NVIDIA H200 GPUs with ray-based data parallelism. As summarized in Table~\ref{tab:runtime_cost}, the platform processed 576 scenarios in 25.48 minutes, corresponding to 6.63 GPU-hours and an average throughput of 11.59 scenarios per minute. Each scenario consists of 11 simulation steps in total, including 3 history steps used for context initialization (Steps 0--2) and 8 planning steps for closed-loop evaluation (Steps 3--10).

\begin{table}[h]
\centering
\caption{\textbf{Runtime cost of the closed-loop platform.}}
\label{tab:runtime_cost}
\begin{tabular}{lcccc}
\toprule
Stage & Scenarios & Wall time (min) & GPU-hours & Time/scenario (s) \\
\midrule
Non-reactive (NR) & 288 & 24.22 & 3.23 & 5.05 \\
Reactive (R)      & 288 & 25.48 & 3.40 & 5.31 \\
Total             & 576 & 25.48$^\dagger$ & 6.63 & --   \\
\bottomrule
\end{tabular}
\par\smallskip
{\footnotesize $^\dagger$ NR and R stages run in parallel; total wall time equals $\max(24.22, 25.48)$}
\end{table}

At the per-step level, each closed-loop planning step comprises three sequential operations: simulator state advancement ($\sim$215--350\,ms, 6--9\%), 3DGS-based sensor rendering ($\sim$490\,ms, 15\%), and end-to-end model inference ($\sim$3,800\,ms, 78\%). The reactive setting incurs a moderate overhead of 5.5\% over the non-reactive setting at the scenario level, primarily attributable to the additional computation in simulator state advancement for reactive traffic generation. Together, these measurements indicate that the closed-loop platform maintains a practical throughput of approximately 11.6 scenarios per minute, confirming its compatibility with large-scale post-training and evaluation.

\subsection{Rendering and Reconstruction Quality of Simulation Engine}
\begin{figure}[h]
    \centering
    \includegraphics[width=\linewidth]{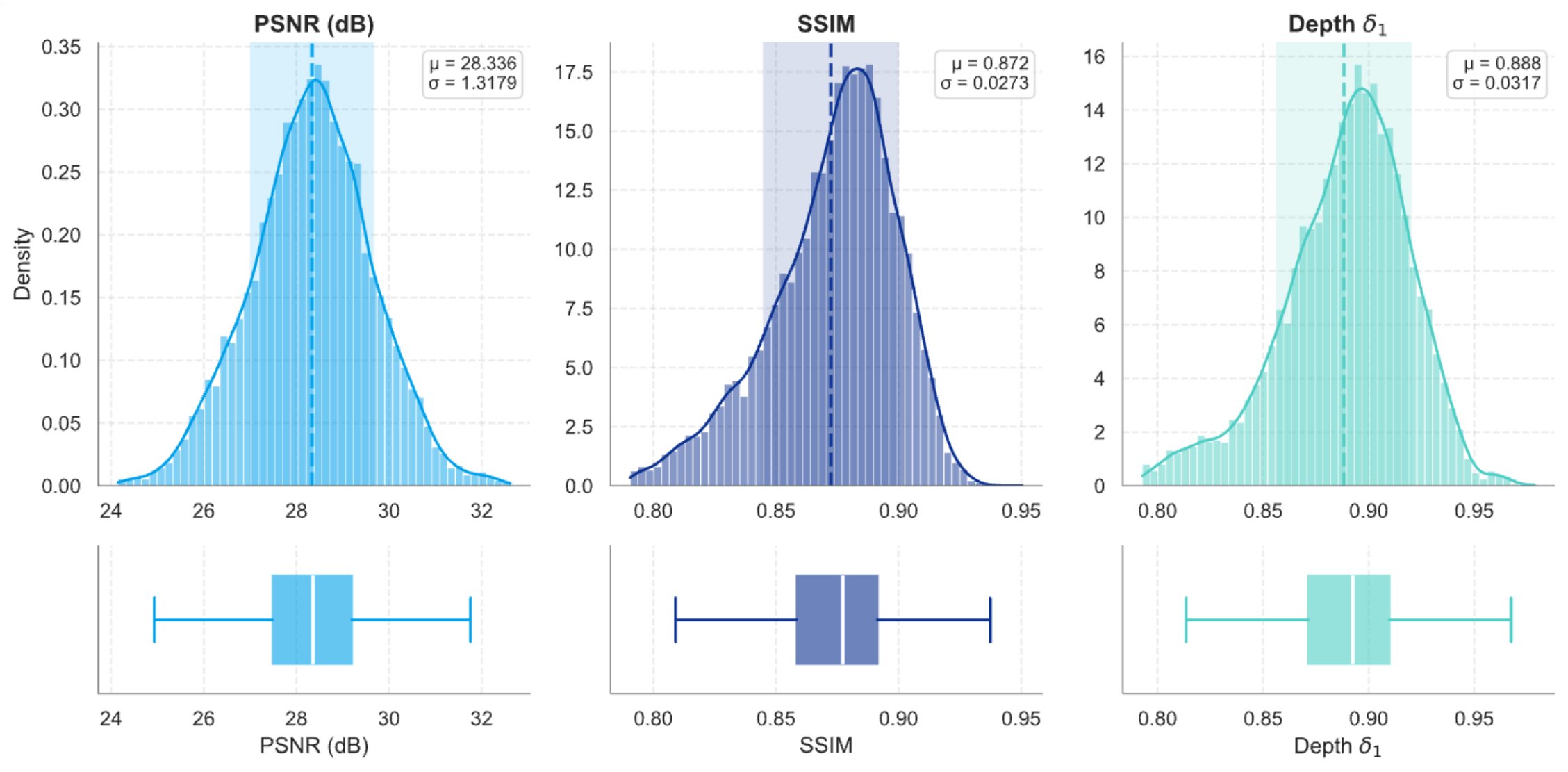}
    \caption{
    \textbf{ Reconstruction quality distribution across multiple metrics.} 
    }
    \label{fig:simulation_engine_distribution}
\end{figure}

We reconstruct 12,862 assets spanning the entire navtrain split to assess the stability of the World Engine reconstruction pipeline, as shown in Fig.~\ref{fig:simulation_engine_distribution}. Image reconstruction quality is evaluated using Peak Signal-to-Noise Ratio (PSNR) and Structural Similarity Index Measure (SSIM), while LiDAR depth quality is assessed using the depth $\delta_1$, defined as the fraction of the predicted depth is within a factor of 1.25 of the ground-truth depth.

\begin{figure}[!th]
    \centering
    \includegraphics[width=0.95\linewidth]{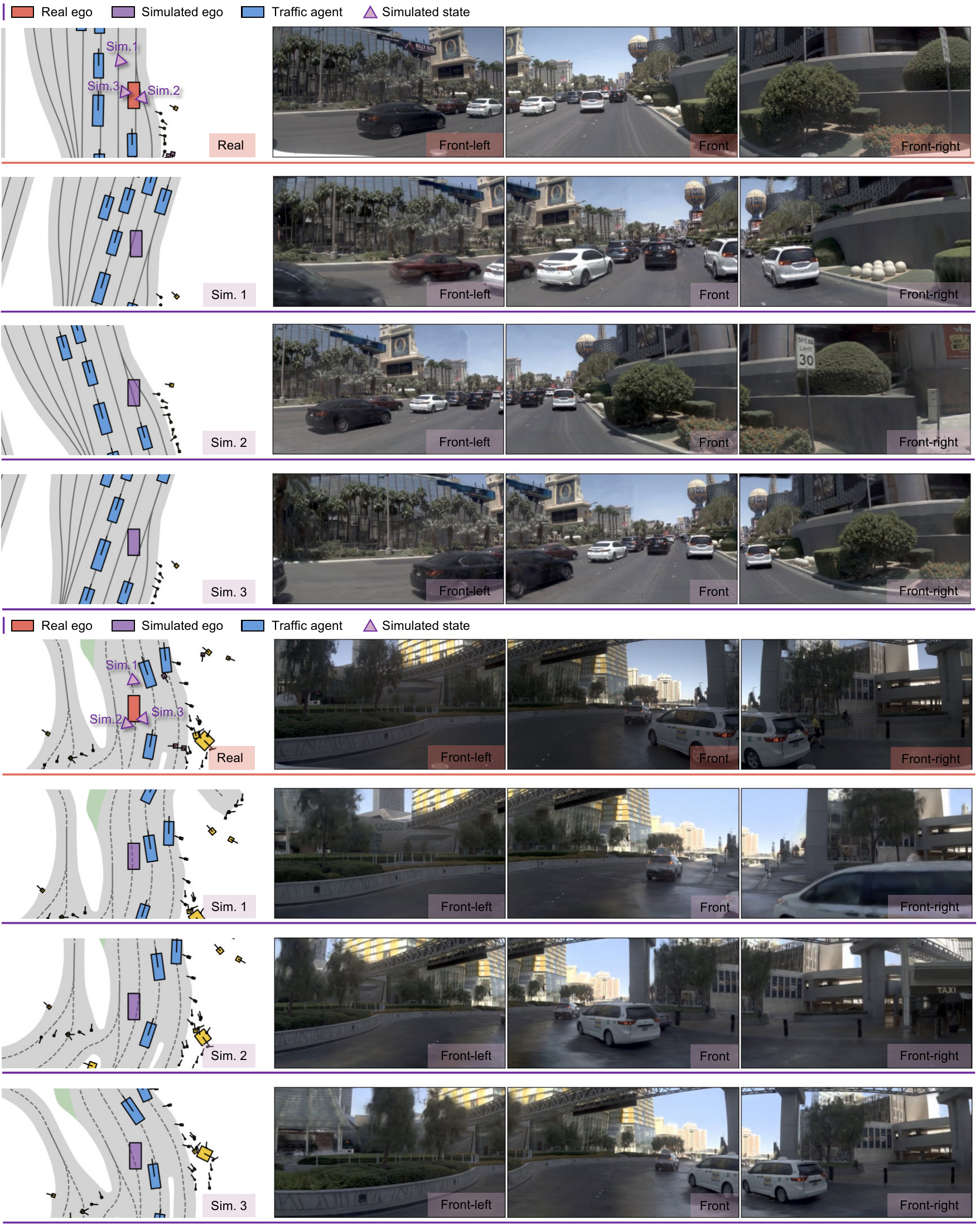}
    \caption{
    \textbf{Visualization of Simulation Engine under simulated states.} ``Sim.'' stands for simulation state.}
    \label{fig:simulation_engine_vis}
\end{figure}

As shown in Fig.~\ref{fig:simulation_engine_vis}, For each reconstructed driving scene, the real logged ego pose and corresponding front-left, front, and front-right camera views are shown in the first row. We then place the ego vehicle at several simulated positions that deviate from the original log trajectory and render the corresponding multi-view observations. The BEV panels show the real ego pose, simulated ego poses, traffic agents, and simulated states. These examples demonstrate that the simulation engine can synthesize realistic camera observations for ego states beyond the logged trajectory, supporting closed-loop simulation and rollout generation.

